**Active Inference Tree Search in Large POMDPs**

Domenico Maisto[1], Francesco Gregoretti[2], Karl Friston[3,4], Giovanni Pezzulo[1,*]

1) Institute of Cognitive Sciences and Technologies, National Research Council, Via Gian Domenico Romagnosi 18/A, Rome 00196, Italy

2) Institute for High Performance Computing and Networking, National Research Council, Via Pietro Castellino 111, Naples 80131, Italy

3) The Wellcome Centre for Human Neuroimaging, Institute of Neurology, University College London, London, WC1N 3AR UK

4) VERSES AI Research Lab, Los Angeles, CA 90016, USA

* Corresponding author:

Giovanni Pezzulo
ISTC-CNR
Via San Martino della Battaglia 44, 00185 Rome, Italy
Email: giovanni.pezzulo@istc.cnr.it
Phone: +39 6 44595206

**Abstract**

The ability to plan ahead efficiently is key for both living organisms and artificial systems. Model-based planning and prospection are widely studied in cognitive neuroscience and artificial intelligence (AI), but from different perspectives—and with different desiderata in mind (biological realism versus scalability) that are difficult to reconcile. Here, we introduce a novel method to plan in POMDPs—*Active Inference Tree Search* (AcT)—that combines the normative character and biological realism of a leading planning theory in neuroscience (Active Inference) and the scalability of tree search methods in AI. This unification enhances both approaches. On the one hand, tree searches enable the biologically grounded, first principle method of active inference to be applied to large-scale problems. On the other hand, active inference provides a principled solution to the exploration–exploitation dilemma, which is often addressed heuristically in tree search methods. Our simulations show that AcT successfully navigates binary trees that are challenging for sampling-based methods, problems that require adaptive exploration, and the large POMDP problem '*RockSample'*—in which AcT reproduces state-of-the-art POMDP solutions. Furthermore, we illustrate how AcT can simulate neurophysiological responses (e.g., in the hippocampus and prefrontal cortex) of humans and other animals that solve large planning problems. These numerical analyses show that *Active Tree Search* is a principled realisation of neuroscientific and AI planning theories, offering biological realism and scalability.

**Keywords**: Active Inference, Tree Search, model-based planning, POMDP

## 1 Introduction

Model-based planning problems have received substantial attention in various disciplines, including AI, machine learning, robotics, cognitive science, and neuroscience. The interdisciplinary exchanges between these disciplines have been numerous [1–6], but yet we still lack a theoretical synthesis that unites their desiderata (e.g., biological realism in neuroscience versus efficiency in AI) [7]. Here, we take a step in this direction by showing that a prevalent theory of model-based control and planning in neuroscience—active inference [8]—can be extended straightforwardly to address large-scale planning problems using tree search methods. Our novel approach—*Active Inference Tree Search (*AcT*)*—bridges computational neuroscience and AI requirements by combining the normative character and biological realism of active inference with the scalability and efficiency of tree search methods.

Active inference is an increasingly popular computational neuroscience framework that characterises perception, action, and planning in terms of approximate (variational) Bayesian inference under a generative model [9–12]. Active inference is related to a family of recent approaches to solving POMDP problems in machine learning and AI, which exploit a general duality between control and inference problems [13] and include *control as inference* [14,15], *planning as inference* [16,17], *risk-sensitive* and *KL control* [18]. A peculiarity of active inference is that it implements a principled form of model-based planning: it infers posteriors over action sequences (or policies) by considering an *expected free energy* functional, effectively balancing exploration and exploitation in a context-sensitive and optimal fashion. Previous studies have established that the computations underlying active inference are biologically plausible and can reproduce various findings in functional brain anatomy, neuronal dynamics and behaviour [11,19–21], as well as furnishing sophisticated forms of inference under hierarchical and temporally deep generative models [12,19,21–25].

However, the active inference framework has been developed with cognitive and biological realism in mind, not scalability or implementational efficiency. Its current implementations require the exhaustive evaluation of all allowable policies and hence can only address small-scale POMDP problems. Here, we develop an extension of active inference aiming to address larger POMDPs: *Active Inference Tree Search* (AcT). Our novel algorithm retains the key aspects of active inference—namely, the use of expected free energy to infer the posterior probability of policies—but relaxes the exhaustive evaluation of all policies, using tree search planning methods that are popular in AI [1–6].

Tree search AI methods perform a look-ahead search over a planning tree, which describes the possible courses of actions and their associated outcome values. The tree is expanded during planning from the root node (i.e., the state where planning starts) to the leaves. In most practical applications, the tree cannot be explored exhaustively. Various heuristic procedures have been proposed to decide what actions to consider next, how to expand the planning tree, and how to balance exploration and exploitation to find an almost-optimal sequential policy [26,27]. A common way to approximate the value of possible policies is using Monte-Carlo sampling [28], which permits sampling rewards obtained by following a given branch of the tree (corresponding to a given course of action) and storing their statistics in the tree nodes. We will show that *Active Inference Tree Search* can contextualise these heuristic methods within a normative and biologically realistic approach, using an (expected) free energy functional that automatically entails the appropriate level of exploration, rendering the use of Monte Carlo methods unnecessary.

The main contribution of this article is a proof of principle that the novel AcT method—that combines active inference and tree search—can augment both approaches. On the one hand, using a planning tree enables active inference to handle larger problems than previously. On the other hand, active inference provides a principled approach to dissolve the exploration-exploitation dilemma, which is addressed heuristically in tree search methods.

In the following Sections, we first review tree search planning methods in AI and active inference. We then introduce *Active Inference Tree Search* formally and validate it using three simulations, showing that it can handle (i) deceptive binary trees (that are challenging for sampling-based methods), (ii) problems that require adaptive exploration, and (iii) larger-scale POMDP problems (i.e., *RockSample*). Finally, to highlight the potential of *Active Inference Tree Search* for studying biological phenomena, we use the scheme to simulate neuronal responses in humans (and other animals) that solve large planning problems.

## 2. Methods: technical background

### *2.1 Partially observed Markov decision processes (POMDP)*

Several real world and biological problems can be cast as sequential decisions under uncertainty. Formally, they can be treated as extensions of Markov Decision Process, where the observed action outcomes provide only partial information about the state of the environment; this corresponds to the framework of Partially Observed Markov Decision Process (POMDP) [5] (Sutton & Barto, 1998).

A POMDP can be defined as a tuple $\langle S, A, T, Z, O, R \rangle$ where:

- $S$ denotes the set of environment states that generate information for the agent to accomplish a task;

- $A$ is the set of actions potentially executable by the agent;

- $T : S \times A \times S \rightarrow [0,1]$, such that $T(s, a, s^{'}) = Pr(s^{'}|s, a)$, is the transition probability of being in a state $s'$ after performing the action $a$ from state $s$;

- $Z$ denotes the set of observations.

- $O : S \times A \times Z \rightarrow [0, 1]$, such that $O(s^{'}, a, z) = Pr(z|a, s^{'})$, is the probability of observing $z \in Z$ in the transited state $s'$ by performing an action $a$;

- $R : S \times A \rightarrow \mathbb{R}$, where $R(s, a)$ is the reward obtained by taking the action $a$ from a particular state $s$ [5].

In a POMDP, the state of the environment cannot be observed directly but can be inferred based on partial observations that the agent solicits through action. Since the agent's state information can be noisy or incomplete, it is helpful for an agent to consider a probability distribution over the states it could be in. This probability distribution, called a (Bayesian) *belief*, is defined as the posterior $b_t(s) = Pr(s_t = s|h_t, b_0)$ given the initial belief $b_0$ and a complete sequence or *history* $h_t = \{a_0, z_1, \dots, z_{t-1}, a_{t-1}, z_t\}$ of past actions and observations. At any time $t$, it is possible to write down the belief state $b_t$ as a Bayesian update $\tau(b_{t-1}, a_{t-1}, z_t)$ of the previous belief state $b_{t-1}$, given the action $a_{t-1}$ and the current observation $z_t$.

Generally, solving a POMDP problem means finding a *plan* or *policy* by predicting the situations the agent could encounter in the future, conditioned on the actions it executes. One can specify a policy as a function $\pi : \mathcal{B} \longrightarrow A$ that associates beliefs $b \in \mathcal{B}$ to actions $a \in A$. In Reinforcement Learning (RL), the value function $V_\pi(b)$ of a policy $\pi$, evaluated in a belief state $b$, corresponds to the expected total discounted reward $V_\pi(b) = \mathbb{E}[\sum_{t=0}^{T} \delta^t R(s_t, \pi(b_t))]$, where $\delta \in (0,1]$ is a discount factor, and T is a finite (or infinite) value if the POMDP problem has a finite (or infinite) time horizon. Following the RL setup, a POMDP plan or policy is optimal when $\pi^*$ maximises the value function $V_{\pi^*}(b)$.

### *2.2 Online POMDP planning*

There are two main approaches to POMDP problems: offline and online. In offline methods [29][30][31][32], the policy is computed before execution by considering every possible belief state. Offline methods achieve good results for small-size scenarios but are not suitable for large POMDP problems. In online methods, there is an alternation between policy construction, whose goal is discovering a good short-path policy (often a single action) for the current belief; and execution, where the selected policy is executed. These methods scale up to large POMDP problems but usually result in suboptimal policies, as they are computed based on a subset of beliefs. The two approaches are complementary and can coexist. For example, online methods can be complemented by initial approximations acquired via some offline algorithm. However, online methods are generally more widely used, given their scalability.

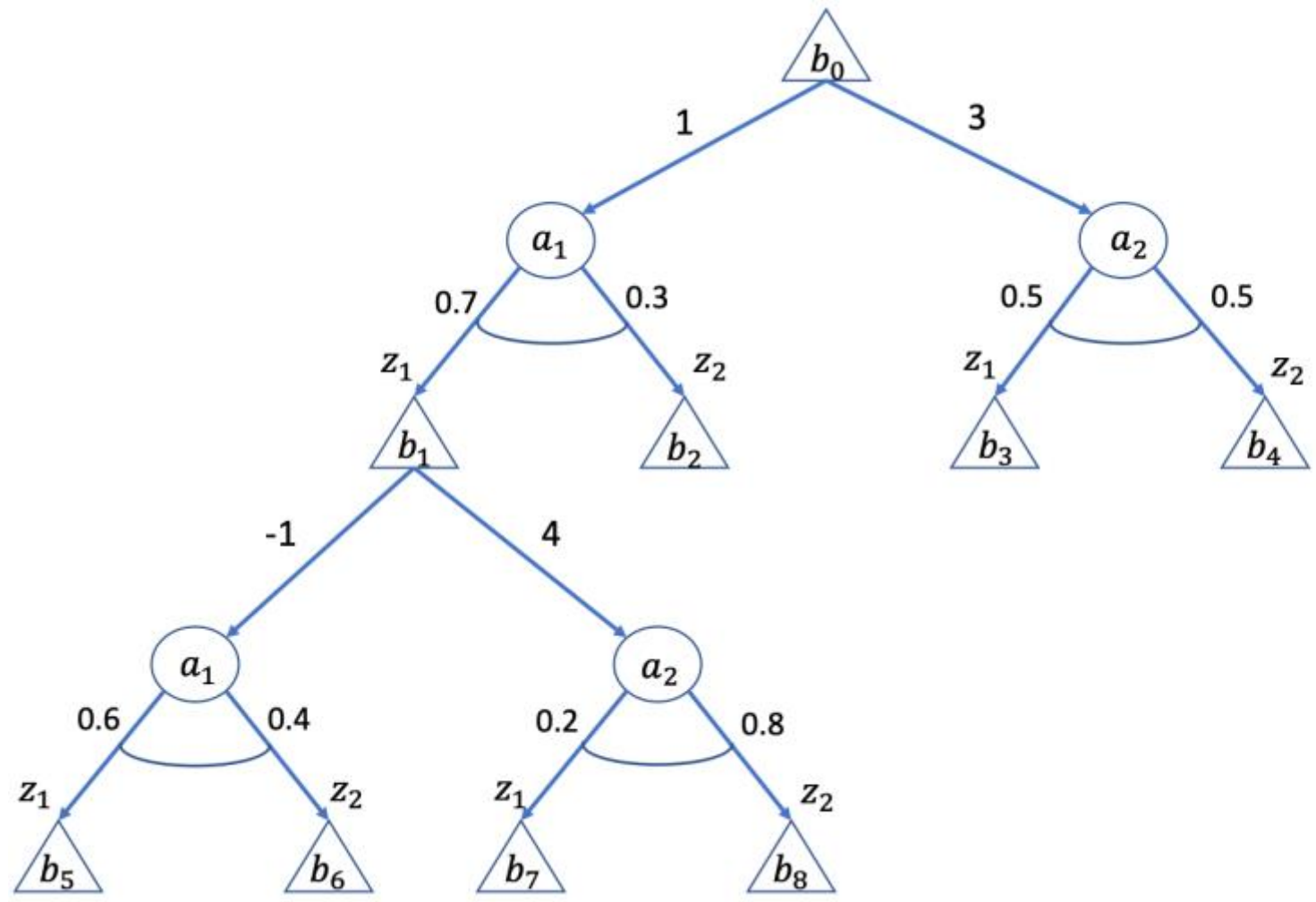

*Figure 1. AND-OR tree for a POMDP with 2 observations $\{z_1, z_2\}$ and 2 actions $\{a_1, a_2\}$. Triangular (OR) nodes represent belief states, whereas circular (AND) nodes represent actions.*

*The numerical values shown on the edges that stem from OR-nodes represent rewards $R(b, a)$, whereas the numerical values shown on the edges that stem from AND-nodes represent conditional probabilities $P(z|b, a)$.*

Ross et al. (Ross, Pineau, Paquet, et al., 2008) established a general scheme for online planning algorithms. In their scheme, every online algorithm can be understood as a procedure in which policy construction is a routine implementing a predefined set of steps: 1) visit, 2) expansion, and 3) estimation of an AND-OR tree, with the OR nodes representing beliefs and AND nodes corresponding to actions (Figure 1). The algorithm starts by setting the current belief as the *root* node of a tree; then builds new belief nodes generated by action nodes. Every time a new belief node is allocated, it is evaluated, and its value is transmitted up to the belief ancestors—up to the root—to update the value of the corresponding policy (i.e., action sequence corresponding to a specific branch of the tree).

The most popular online-planning approaches are Heuristic Search, Branch-and-Bound pruning, and Monte Carlo sampling [33]. In Heuristic Search methods [34][35], a routine explores the belief tree using a heuristic to detect relevant nodes to branch out (frequently, for a single forward step). It successively updates the heuristic value associated with its ancestors (which differs between heuristic search algorithms). However, this procedure can be computationally expensive, reducing the effectiveness of the heuristic-based node selection. Branch-and-bound approaches instead rely on a general search technique that constrains the search tree expansion by pruning suboptimal branches [36]. They assign every belief tree node an upper and a lower bound of a quality value function. If a branch leads to a node with an upper bound—that is lower than the lower bound of another node of a different branch—then the first node is labelled as the root of a suboptimal subtree that can be pruned. Finally, Monte Carlo algorithms randomly sample a subset of observations each time. This procedure constrains the branching expansion of the belief tree and the depth of the search.

While Monte Carlo algorithms follow the same general procedure, they sample outcomes in different ways [37][38]. One of the earliest Monte-Carlo-sampling-based algorithms for solving POMDP—the Sparse Sampling algorithm of Kearns et al. [27]—builds a fixed depth tree search in one stage (i.e., from the root to the leaves) using a "black-box" simulator (a generative model) for modelling state transitions and simulating reward returns. To improve the performance of the Sparse Sampling algorithm, Kocsis and Szepesvári proposed the Upper Confidence Tree (UCT) algorithm [26], which introduced two essential novelties. First, it uses Monte Carlo Tree Search (MCTS) [39]: a rollout-based Monte Carlo planning method inspired by game strategy searches but builds the belief tree progressively and iteratively. Second, it selects actions during the planning phase in a stochastic way rather than by drawing from a uniform distribution ( consistent with theoretical results on sequential decision making under uncertainty [40]).

### *2.3 Related works*

UCT [39] is an online decision-making algorithm that works by constructing a tree of simulated histories $h_t$ that expands from an initial belief state $b_0$ cast as root, and alternating state and action nodes, eventually drawn by a *generative model*, i.e., a probabilistic model that statistically describes the POMDP distributions $T$ and $Z$. To select which node (and branch corresponding to some history $h_t$) to expand next in $h_t a = \{a_0, z_1, \dots, z_t, a_{t+1}\}$, the algorithm uses the value function $V_\pi(h_t a)$ and evaluates the expected return from the initial belief $b_0$ following the policy $\pi$ furnished by $h$. A peculiarity of UCT, as of every MCTS algorithm, is the way the value function $V_\pi(h_t a)$ is calculated. Instead of bootstrapping $V_\pi(h_t a)$ using dynamic programming, it is estimated by Monte Carlo sampling, where multiple stochastic rollouts approximate a mean value. The computed value is then propagated back to each branch node and averaged with contributions from other histories branching off from the same node. Concurrently, a visitation count $N(h_t a)$ is updated, such that $N(h_t) = \sum_a N(h_t a)$ is the number of simulations ran through the node representing $s$. In UCT, node visit counts are used effectively: node selection is seen as a Multi-armed Bandit problem for which the optimal choice uses the Upper Confidence Bound (UCB) $V_\pi(h_t a) + c_p \sqrt{\log N(h_t) / N(h_t a)}$. UCB augments the value function with an exploration term favouring less-visited nodes [40].

A UCT-based planning algorithm that has received attention in the last decade is POMCP [41]. POMCP can handle POMDPs with large state spaces. It adopts MCTS to generate an AND-OR belief tree, where the AND nodes (actions) are selected through the UCB algorithm, and the OR nodes represent a set of sampled states (not a full probability distribution), which are iteratively maintained by a particle filter. Although it can handle problems of considerable size, POMCP has some implicit limitations. By representing the POMDP problem as a belief tree, POMCP needs to visit every potential observation related to a belief state at least once. Furthermore, as it uses the UCB heuristic, its worst-case is computationally challenging [42].

DESPOT [43] is another state-of-art MCTS-based algorithm that tries to overcome (at least theoretically) the above limitations by operating on a sparse belief tree generated on a subset of sampled observations. As in POMCP, the nodes of such a reduced tree—called DESPOT tree—approximate distributions over belief states using particles. An MCTS planning routine progressively constructs the DESPOT tree by iterating the following three stages: a forward search that traverses the tree until it encounters a leaf node according to the heuristic values (which includes a pre-computed regularisation term to prevent overfitting); a leaf initialisation, where a Monte Carlo sampling estimates the upper and lower bounds of the selected leaf node; and a backup, that passes back through the path tracked in the forward search and updates the upper and lower bounds of each visited node, according to the Bellman optimality principle. These three stages are analogous to the selection, expansion, and backpropagation phases of POMCP. However, they are iterated until the difference between the upper and lower bounds of the belief root state is sufficiently small (as in a Branch-and-Bound method).

More recently, new approaches to online POMDP planning allow the parallelisation of extant methods [44][45] or use deep learning to extract and aggregate relevant information from the environment to speed up and improve policy inference [46][47]. Furthermore, applied research in decision making for autonomous urban vehicles has engendered novel solutions to the online POMDP and approximate solutions [48][49][50].

### *2.4 Active inference*

Active inference integrates cybernetic feedback and error control concepts with Bayesian inference [51] with a Bayesian inferential scheme [19,52,53]. It involves a closed-loop process where perception and action selection operate through approximate Bayesian inference [17,54–56], employing a variational approximation under the free-energy-minimisation principle (Friston et al., 2012). The scheme has various proposed variants, and its biological plausibility is under investigation (Friston et al., 2017; Friston et al., 2016a)

Essentially, active inference is a theory of decision-making under uncertainty, positing that agents minimise the expected free energy of future outcomes [10]. It aligns with optimal decision theory and can be viewed as a partially observable Markov decision process (POMDP), wherein the agent holds beliefs about the probability of associating observations with hidden states, with rewards (or cost functions) absorbed into beliefs about initial state distribution and terminal observations [57].

In this setting, active inference can be formally represented by a tuple $\langle S, O, U, \gamma, R, P, Q \rangle$ where:

- $S$ is the set of agent's hidden states $s$ by which the agent infers the environmental state. Where a sequence of hidden states is denoted by $\tilde{s} = (s_0, \dots, s_T)$;
- $O$ is a finite set of outcomes (i.e., observations from the environment) $o$, and $\tilde{o} = (o_0, \dots, o_T)$;
- $U$ is a finite set of *control states* $u$ executable by the agent to control the environment. A sequence of *control states* $\tilde{u} = [u_t, \dots, u_T]$ is called *policy* and denoted as $\pi$. Thus, $\pi = \tilde{u} = [\pi^{(t)}, \dots, \pi^{(T)}]$;
- $\gamma \in \mathbb{R}$, is an additional variable denoted as *precision*, introduced to self-tune the control-state selection process adaptively;
- $R(\tilde{o}, \tilde{s}, \tilde{a})$ is a *generative process* that generates probabilistic sequences of outcomes from hidden states and actions $\tilde{a} = (a_0, \dots, a_T)$ corresponding to the activations of the control states $\tilde{u}$ in the environment.
- $P(\tilde{o}, \tilde{s}, \pi, \gamma | \Theta)$ is a *generative model* with parameters $\Theta = \{\mathbf{A}, \mathbf{B}, \mathbf{C}, \mathbf{D}, \mathbf{E}, \alpha, \beta\}$ (defined later on), over outcomes, hidden states, control states and precision;
- $Q(\tilde{s}, \pi, \gamma)$ is an approximate posterior distribution over states, control states, and precision, with expectations $(\boldsymbol{s}_0^{\pi}, \dots, \boldsymbol{s}_T^{\pi}, \boldsymbol{\pi}, \boldsymbol{\gamma})$, approximating the posterior of the generative model $P(\tilde{s}, \tilde{u}, \gamma | \tilde{o}, \Theta)$.

It is worth noting that the *generative process* describes transitions in the environment due to the agent's actions, generating observed outcomes. In contrast, the *generative model* reflects the agent's beliefs, encoding states and policies as expectations. Notably, there is a distinction between control states $u$ in the generative model and actions $a$ in the generative process. Action is a real variable that acts on the environment, while the corresponding hidden cause in the generative model is a control state. This means the agent needs to infer its behaviour by forming beliefs about control states using the observed consequences of its action. That allows for the formulation of action in terms of beliefs about policies and transforms an optimal control problem into an optimal inference problem, known as planning as inference (Attias, 2003).

### *2.4.1 Generative models for active inference*

As shown in Fig. 2, the generative model used in active inference includes hidden *states* $s$ as causes of the observed outcomes $o$. Hidden states move forward in time under a policy $\pi$ that depends on the precision $\gamma$. A series of factorisations permits writing down the model's joint density as:

$$P(\tilde{o},\tilde{s},\tilde{u},\gamma|\Theta) = P(\gamma|\Theta)P(\pi|\gamma,\Theta)\prod_{t=0}^{T} P(o_t|s_t,\Theta)P(s_t|s_{t-1},\pi,\Theta) \tag{1}$$

where:

$$P(o_t\,|s_t,\Theta) = \mathbf{A}$$

$$P(s_{t+1}|s_t,\pi,\Theta) = \mathbf{B}\big(u_t = \pi^{(t)}\big) \text{ for } t > 0 \text{ and } P(s_0|\Theta) = \mathbf{D} \text{ otherwise} \tag{2}$$

$$P(o_t|\Theta) = \mathbf{C}$$

$$P(\pi|\gamma,\Theta) = \sigma(\ln \mathbf{E} - \gamma \cdot \mathbf{G}_\pi|\Theta)$$

$$P(\gamma|\Theta) \sim \Gamma(\alpha,\beta)$$

In Equations (2), the matrix $\mathbf{A}$ (with size $|O| \times |S|$) encodes the likelihood of observations given a hidden state, while $\mathbf{C}$, a vector of size $|O|$, represents their prior distribution. Importantly, the prior distribution of observations, encoded in the vector $\mathbf{C}$, specifies the agent's *preferred* observations, which play an analogous role to rewards in Reinforcement Learning and can be defined in a task-dependent manner. The matrices $\mathbf{B}(u_t)$ (of size $|S| \times |S|$) define $u$-specific state transitions, while $\mathbf{D}$ is a $|S|$-long vector encoding the prior distribution of the initial state, and $\mathbf{E}$ is the prior expectation of each policy that can be specified according to the problem at hand. Finally, $\sigma$ is the Boltzmann distribution and $\alpha$ and $\beta$ correspond to the shape and the rate parameters of the gamma density, which underlies the $\gamma$-distribution, respectively.

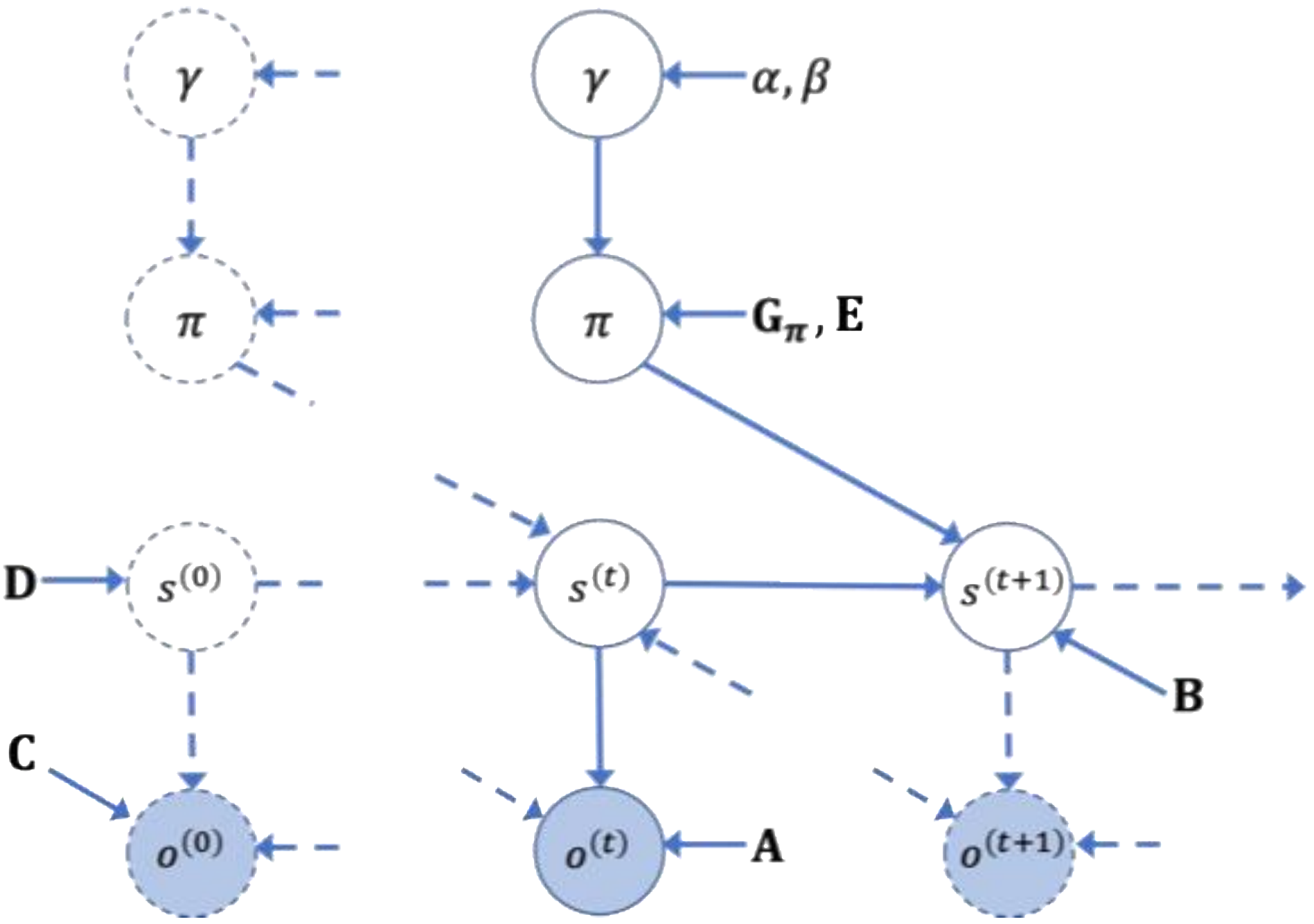

*Figure 2. Graphical model for active inference. See the main text for an explanation.*

The quantity $\mathbf{G}_\pi$ is a score of the "quality" of a generic policy. It can be viewed as the log prior of a given policy, conditioned on the future state and observations, together with preferred outcomes (see below for more details).

### *2.4.2 Approximate Bayesian inference and variational free energy minimisation*

Perception and action in Active Inference are read as predictive processes, in which an agent infers the states of the external world. Active inference is an application of the free energy principle; namely a variational principle in which the (path integral) of free energy is minimised. The free energy in question is an upper bound on the negative log marginal likelihood of sensory samples. This means that — under active inference — action and perception minimise the self-information (a.k.a., surprisal or surprise) of sensory data. This is equivalent to maximising the Bayesian model evidence of a generative model of those data. In terms of minimising surprise, agents continuously compare predictions (from the generative model) and sensory stimuli (from the generative process). Discrepancies between sensations and predictions of those sensations are minimised either by modifying predictions (i.e., perception) or acting to change the world and subsequent sensory samples (i.e., action).

In most cases, exact Bayesian inference cannot be realised because the computation of model evidence $P(\tilde{o}|\Theta)$ and requisite posterior $P(\tilde{s}, \pi, \gamma|\tilde{o}, \Theta)$ and are (generally) intractable. Active Inference therefore appeals to approximate Bayesian inference in which intractable inference problem is transformed into an optimization problem by substituting $P(\tilde{s}, \pi, \gamma|\tilde{o}, \Theta)$ and $P(\tilde{o}|\Theta)$ with two quantities — an approximate posterior and a variational free energy — that are computable [58]. Under this variational formulation, the optimality principle underlying Active inference becomes the minimisation of variational free energy, defined as follows:

$$
\begin{aligned}
F(\tilde{o}) \quad &= \mathbb{E}_{Q}[\ln Q(\tilde{s},\pi,\gamma) - \ln P(\tilde{o},\tilde{s},\pi,\gamma|\Theta)] \\
&= D_{\mathrm{KL}}[Q(\tilde{s},\pi,\gamma)||P(\tilde{s},\pi,\gamma|\tilde{o},\Theta)] - \ln P(\tilde{o}|\Theta) \\
&\geq \; -\ln P(\tilde{o}|\Theta)
\end{aligned} \tag{3}
$$

where $\mathrm{E_Q}[\cdot]$ denotes an expectation under the approximate posterior Q, $\mathrm{D_{KL}}[\cdot\,||\,\cdot]$ is the Kullback-Leibler divergence, and $-\ln \mathrm{P}(\Theta)$ (i.e., the negative logarithm of the model evidence $\mathrm{P}(\Theta)$) is called self-information, surprisal or, more simply, surprise. Equation (3) implies that one needs to minimise the variational free energy to approximate the model evidence[1]. The approximate posterior is equipped with a known functional form, which generally factorises over hidden states and parameters $(\tilde{\mathrm{s}},\pi,\gamma)$:

$$
Q(\tilde{s},\pi,\gamma) = \; Q(\pi)Q(\gamma)\prod_{t=0}^{T} Q(s_t|\pi); \tag{4}
$$

This factorisation is known as a mean-field approximation. When $Q(\tilde{s},\pi,\gamma)$ converges on the posterior $P(\tilde{s},\pi,\gamma|\tilde{o},\Theta)$, the variational free energy decreases. If they match exactly, and their divergence is zero, free energy becomes the surprise. Therefore, one could summarise variational inference [58] as minimising free energy to approximate the posterior $P$ with $Q$ while, at the same time, evaluating a bound on the log-model evidence $\ln P(o|\Theta)$ (often called an evidence bound in machine learning). Effectively, this converts an intractable marginalisation problem into an optimisation problem with easy-to-compute approximations to log evidence (a.k.a., approximate Bayesian inference).

It is possible to demonstrate [8] that the optimal solution that minimises $F(\tilde{o},\tilde{s},\pi,\gamma)$ is represented by the approximate posterior $Q$ with sufficient statistics $\boldsymbol{\mu} = (\tilde{\boldsymbol{s}}^{\pi},\boldsymbol{\pi},\boldsymbol{\gamma})$, with $\tilde{\boldsymbol{s}}^{\pi} = \boldsymbol{s}_0^{\pi},\ldots,\boldsymbol{s}_{\boldsymbol{T}}^{\boldsymbol{\pi}}$ set, at any time $t$, as:

$$
\boldsymbol{s}_t^{\pi} \approx \begin{cases} \sigma\left(\ln \mathbf{A}\cdot \mathrm{o}_t + \ln\left(\mathbf{B}\left(\pi^{(t-1)}\right)\cdot \boldsymbol{s}_{t-1}^{\pi}\right)\right) & \text{if} \quad t>1 \\ \sigma(\ln \mathbf{A}\cdot \mathrm{o}_t + \ln \mathbf{D}) & \text{if} \quad t=1 \end{cases}
$$

$$
\boldsymbol{\pi} = \sigma(\ln\mathbf{E} - \; \boldsymbol{\gamma}\cdot \mathbf{G}_{\pi}) \tag{5}
$$

$$
\boldsymbol{\gamma} = \frac{\alpha}{\beta - \mathbf{G}_{\pi}}
$$

Here, we use the symbol "$\cdot$" to denote the inner product, defined as $\mathbf{A}\cdot\mathbf{B} = \mathbf{A}^T\mathbf{B}$, where A and B are two arbitrary matrices. The first equation defines the expected hidden state and corresponds to the part of active inference that implements perception. For $t = 1$, by considering Equation (2), we have $\boldsymbol{s}_t^{\pi} \approx \sigma(\ln \mathbf{A}\cdot \mathrm{o}_t + \ln \mathbf{D})$.

[1] Generally, policies are treated as sequences of actions from the current time point, where realised actions — taken in the past — are absorbed into observations when necessary.

The hidden state $\boldsymbol{s}_t^{\pi}$ estimated at a certain time point depends both on the expected hidden state at the previous time and the actual outcome $o_t$ observed after executing the action predicted by $\pi^{(t-1)}$. The second equation derives (as the expected hidden state) from a Boltzmann distribution of the policies' quality values. The expected value of $\gamma$ is the distribution's sensitivity (or inverse temperature parameter): it adjusts the tendency to select a policy with greater or lesser confidence. The last equation tunes the expected precision value on the base of policy quality values (in a nonbiological setting, this precision is usually set to 1, especially for policies that only look ahead).

The term $\mathbf{G}_{\pi}$ is the policies' *expected free energy* (EFE). It is used to score the quality of a generic policy with respect to the future outcomes and states that are expected under such policies. It controls the optimism bias in the first equation, determines policy selection in the second equation, and nuances an agent's confidence in action selection in the third equation. With greater differences among the values of $\mathbf{G}_{\pi}$, the precision is greater—and the agent is more confident about what to do next (in a non-biological setting, the selected policy is the one with the smallest EFE).

Note a fundamental difference between active inference and RL based approaches to POMDP: RL approaches are based upon a value *function* of future states, while active inference infers the best policy using an expected free energy *functional of beliefs* about future states. Technically, this means replacing the Bellman optimality principle with a straightforward principle of least action, where the action is the path integral of expected free energy. Teleologically, this means active inference considers optimal sequences of belief states that subsume information-seeking and preference-seeking imperatives into the same functional, thereby dissolving the exploration–exploitation dilemma.

To get a closed form for $\mathbf{G}_{\pi}$ we need to premise that the Evidence Lower BOund – the ELBO – of the model evidence $P(o|\pi)$ conditioned by the policy $\pi$ corresponds to the negative variational free energy $-F(o,\pi)$. Using the Jensen inequality, we can write down:

$$\log P(o|\pi) = \log \mathbb{E}_{\mathrm{Q}(s|\pi)}\left[\frac{P(s,o|\pi)}{Q(s|\pi)}\right] \geq \mathbb{E}_{\mathrm{Q}(s|\pi)}\left[\log \frac{P(s,o|\pi)}{Q(s|\pi)}\right] = -F(o,\pi)$$

Considering that EFE is a variational free energy of future outcomes and future states under a fixed policy $\pi$, from the current instant $t$ to some horizon $T$, we can define $\mathbf{G}_{\pi}$ (in analogy with the definition of variational free energy given in Equation (3)) as:

$$\mathbf{G}_{\pi} = \sum_{\tau=t}^{T} \mathrm{G}(\pi,\tau) = \sum_{\tau=t}^{T} \mathbb{E}_{Q(s_\tau,o_\tau|\pi)}\left[\ln \frac{Q(s_\tau|\pi)}{P(s_\tau,o_\tau|\pi,\mathbf{C})}\right] \tag{6}$$

where

$$\begin{aligned} \mathrm{G}(\pi,\tau) &= \mathbb{E}_{\tilde{Q}}[\ln Q(s_\tau|\pi) - \ln P(s_\tau,o_\tau|\pi,\mathbf{C})] \\ &= \mathbb{E}_{\tilde{Q}}[\ln Q(s_\tau|\pi) - \ln P(s_\tau|o_\tau,\pi) - \ln P(o_\tau|\mathbf{C})] \end{aligned} \tag{7}$$

$$
\begin{aligned}
&\approx \mathbb{E}_{\tilde{Q}}[\ln Q(s_\tau|\pi) - \ln Q(s_\tau|o_\tau,\pi)] - \mathbb{E}_{\tilde{Q}}[\ln P(o_\tau|\mathbf{C})] \\
&= \mathbb{E}_{\tilde{Q}}[\ln Q(o_\tau|\pi) - \ln Q(o_\tau|s_\tau,\pi)] - \mathbb{E}_{\tilde{Q}}[\ln P(o_\tau|\mathbf{C})] \\
&= D_{\mathrm{KL}}[Q(o_\tau|\pi)||P(o_\tau|\mathbf{C})] + \mathbb{E}_{\tilde{Q}}\big[H[P(o_\tau|s_\tau)]\big].
\end{aligned}
$$

Here, $P(s_\tau, o_\tau|\pi, \mathbf{C}) \triangleq P(o_\tau|\mathbf{C})P(s_\tau|o_\tau,\pi)$ and $\mathbb{E}_{\tilde{Q}}[\cdot]$ is the expected value under the predicted posterior distribution $\tilde{Q} = Q(o_\tau, s_\tau|\pi) \triangleq P(o_\tau|s_\tau)Q(s_\tau|\pi)$ over hidden states and their outcomes under a specific policy $\pi$. The final identity in Equation (7) provides an interpretation of the expected free energy as a sum of two terms. The former is the Kullback-Leibler divergence between (approximate) posterior and prior over the outcomes; it constitutes the quality score's *pragmatic* (or utility-maximising) component, favouring policies that realise expected outcomes under the generative model. The latter is the expected entropy under the posterior over hidden states; it represents the quality score's *epistemic* (or ambiguity-minimising) component, favouring policies that realise unambiguous outcomes. In other words, the former (pragmatic) term represents the *risk* that the anticipated outcomes $Q(o_\tau|\pi)$ diverge from prior preferences $P(o_\tau)$, while the latter (epistemic) minimises *ambiguity*. In summary, risk measures the difference between predicted outcomes $\boldsymbol{o}_\tau^\pi$ in the future and preferred outcomes encoded in the **C** vector, while ambiguity quantifies to what extent a future state $\boldsymbol{s}_\tau^\pi$ diminishes uncertainty about future outcomes. From a machine learning perspective, this would be equivalent to saying that $\mathbf{G}_\pi$ embodies a "regularisation" term, which balances between exploitive (pragmatic) and exploratory (epistemic) behaviour.

The expected free energy of a policy $\mathbf{G}_\pi$ can be expressed in terms of linear algebra by considering the equations for the free energy minimising sufficient statistics above, together with the generative model:

$$
\mathrm{G}(\pi,\tau) = \boldsymbol{o}_\tau^\pi \cdot (\ln \boldsymbol{o}_\tau^\pi - \ln P(o_\tau)) + \boldsymbol{s}_\tau^\pi \cdot \mathbf{H} \tag{8}
$$

with

$$
\begin{aligned}
&\boldsymbol{s}_\tau^\pi = \mathbf{B}\big(u_\tau = \pi^{(\tau)}\big) \cdot \boldsymbol{s}_{\tau-1}^\pi \\
&\boldsymbol{o}_\tau^\pi = \mathbf{A} \cdot \boldsymbol{s}_\tau^\pi \\
&\ln P(o_\tau) = \ln \mathbf{C} \\
&\mathbf{H} = -\mathrm{diag}(\mathbf{A} \cdot \ln \mathbf{A})
\end{aligned} \tag{9}
$$

where $\ln P(o_\tau)$ is the log-vector of preferred outcomes, and **H** is the entropy matrix pertaining to future outcomes. This provides a convenient way to evaluate a policy's expected free energy.

Planning in Active Inference entails simulating the future, using Equation (5) and posterior beliefs about the hidden state $\boldsymbol{s}_t^\pi$ at the current time and computing—for each policy $\pi^{(\tau)}$, with $\tau > t$—the expected free energy $G(\pi,\tau)$ of Equation (8). In other words, expected free energy is evaluated using the predictive posterior over future states (and outcomes). This means the predicted observations $\boldsymbol{o}_\tau^\pi$ used for planning are not actual observations; rather, they are random variables in the future, with a predictive posterior for each policy. This calls for an expectation or averaging; hence, expected free energy.

Crucially, standard implementations of active inference (with a few exceptions [59–62]) assume that an agent evaluates all policies $\pi$ for any possible future state an agent could be in. This entails computing the expected free energy $\boldsymbol{G}_{\pi}$ of each policy $\pi$. Once every plausible (i.e., allowable) policy has been scored (and its associated "quality" evaluated), the agent uses a Boltzmann distribution—defined by the last two expressions of Equation (5)—to form posterior beliefs over policies, from which the next action is selected for execution (by sampling from the posterior distribution).This approach has been used to address a variety of cognitive phenomena, including decision-making [20,63], habitual behaviour, salience, and curiosity-driven planning [64–67], and to develop a process theory for neural computation [11,68]. However, the necessity to evaluate all policies exhaustively renders the approach unable to solve large POMDP problems. To fill this gap, we introduce Active Inference Tree Search below.

### 3. Active Inference Tree Search

A straightforward method to overcome the limitations of active inference in large POMDPs is to elude the exhaustive evaluation of all allowable policies using a heuristic procedure [69] [24]. The efficacy of such a proposal needs to be assessed, considering the quality of the approximation and the gains in terms of computational costs and tractability. Here, we develop and evaluate a novel tree search scheme to render active inference in large POMDPs tractable: namely, Active Inference Tree Search (AcT).

AcT solves POMDP problems using a planning tree: an abstract structure in which nodes $v_{\tau}$ correspond to beliefs $\boldsymbol{s}_{\tau}^{\pi}$ about the states that can be reached from the current state and where branches represent possible actions that can be taken to reach future states. The planning tree used in AcT differs from the tree commonly used in Reinforcement Learning. The tree is a structure that brings together all the instances of the POMDP problem generated by applying a sequence of possible actions (the policies) to a single instance (the root). Consequently, every path from the root to a leaf corresponds to a potential evolution of the POMDP problem, given a specific policy.

The goal of using a planning tree is estimating the posterior over control states $P(u)$ from which the best action $a_t$ can be sampled. This estimate is obtained through simulations, which start from the current state $s_t$ (with a related observation $o_t$) and proceed forward through specific branches of the decision tree, corresponding to a series of paths or histories $h_{\tau} = (v_t u_t, v_{t+1} u_{t+1}, \cdots, v_{\tau})$. Here, we use $\tau$ to indicate the planning time steps, whereas we use $t$ to indicate root quantities; namely, universal or clock time in the environment. The simulations approximate, statistically, the expected free energy values G of the policies $\pi \equiv (u_t, u_{t+1}, u_{t+2}, \ldots, u_{\tau-1})$; the larger the number of simulations, the more reliable the approximation of G. This planning process is iterated until one or more halting conditions are satisfied. The depth $d$ of the planning tree depends on two control parameters: the discount factor $\delta$, and the discount horizon $\varepsilon$. In our simulations, the maximum depth of a planning tree—and consequently the maximum number of simulations employed to generate it—is fixed by imposing $\delta^{d} < \varepsilon$.

Note that AcT is an algorithm to build a planning tree, not to select actions. Here, we assume that after the planning tree has been built, the agent selects an action by sampling from the distribution of control states inferred at the root node and executes it. At this point, the agent makes a transition to a new state and receives a new observation—and can start planning again.

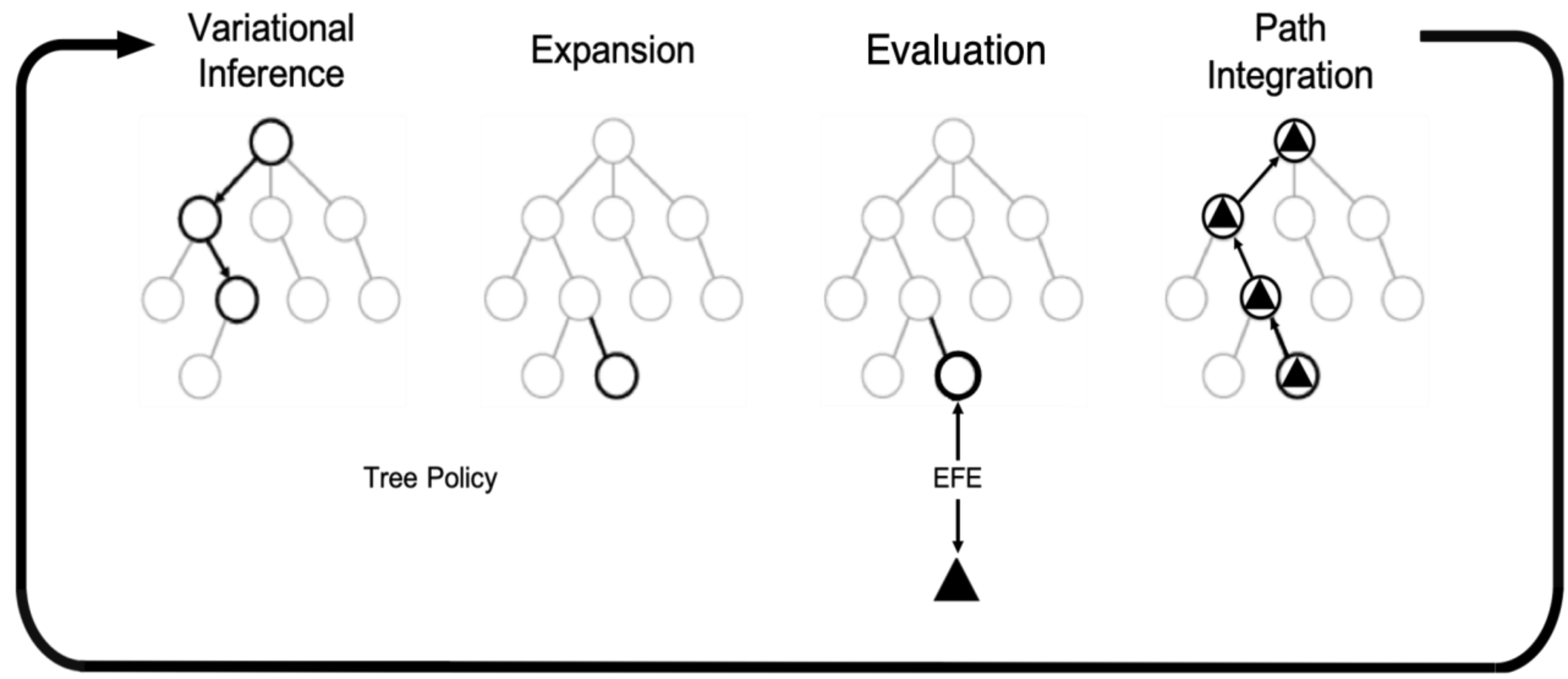

*Figure 3. The four stages of the Active Inference Tree Search (AcT) algorithm. Note that the first two stages (variational inference and expansion) could be considered as two parts of a single TreePolicy procedure of MCTS. The third (evaluation) and the fourth (path integration) stages could be called Eval and PahtIntegration procedures, respectively. The third (evaluation) stage can also be called an Eval procedure. See the main text for an illustration of each of the four stages.*

### *3.1 The four stages of Active Inference Tree Search*

Active Inference Tree Search comprises four successive stages—Variational Inference, Expansion, Evaluation, and Path Integration—applied iteratively at each time step $t$ (Figure 3) for a time $\tau > t$ depending on the exit condition related to a discount factor $\delta$, widely adopted in Monte Carlo Tree Search [26][70]. In the following, we examine each stage in detail. Furthermore, in Algorithm 1, we report the pseudocode of the AcT algorithm (with subroutines) as a function of the parameters $(\mathbf{A}, \mathbf{B}, \mathbf{C}, \mathbf{D})$ commonly adopted in active inference [10][11] and the aforementioned discount factor.

Please note that from now on, we use $\mathbf{x}_\tau$ rather than $\boldsymbol{s}_\tau^\pi$ to denote beliefs, to avoid confusions with the actual states $s_t$.

#### *3.1.1 First stage: Variational Inference*

The goal of the first (variational inference) stage (implemented in VARIATIONALINFERENCE of Algorithm 1) is to select the next non-terminal leaf node $v_\tau$ of the tree to expand. From the root—and recursively until reaching an expandable node of the planning tree—, by using the second one of Equations (5), this stage samples an action over a Boltzmann distribution $\sigma\left(\kappa_p \ln \mathbf{E} - \gamma_\tau \cdot \mathrm{G}(v_\tau)\right)$ that depends on three terms: 1) the EFE $\mathrm{G}(v_\tau) = (\mathbf{A} \cdot \mathbf{x}_\tau) \cdot (\ln(\mathbf{A} \cdot \mathbf{x}_\tau) - \ln \mathbf{C}) - \mathbf{x}_\tau \cdot \left(\sum_i \mathbf{A}_{ij} \ln \mathbf{A}_{ij}\right)_j$, defined in Equations (8) and (9), of the policy $\pi_\tau$ assembled so far, 2) the precision $\gamma_\tau$ computed at each depth of the tree visit, and 3) the prior belief about the policy $\mathbf{E}$ [71].

**Algorithm 1.** Active Inference Tree Search algorithm with subroutines

**function** AcT($\mathbf{A}, \mathbf{B}, \mathbf{C}, \mathbf{D}, \delta$)
  $t \longleftarrow 0$
  **while** halting conditions are not satisfied **do**
    $\mathbf{x}_t \longleftarrow$ update expected state belief $\mathbf{x}_{t-1}$ by using the first line of Equation (5)
    create a node $v(\mathbf{x}_t, u_{t-1})$
    **while** $\delta^\tau < \varepsilon$ **do**
      $v_\tau \longleftarrow$TREEPOLICY$(v, \mathbf{B})$
      $\mathrm{G}_\Delta \longleftarrow$EVAL$(v_\tau, \mathbf{A}, \mathbf{B}, \mathbf{C}, \delta)$
      PATHINTEGRATION($v_\tau$, $\mathrm{G}_\Delta$)
    **End**
    $(s_{t+1}, o_{t+1}, \mathbf{x}_t, a_t) \longleftarrow$extract information saved in $v$
    $t \longleftarrow t + 1$
  **end**
**end**

**function** TREEPOLICY$(v)$
  **while** $v$ is nonterminal **do**
    **if** $v$ not fully expanded **then**
      **return** EXPANSION$(v, \mathbf{B})$
    **else**
      $v' \longleftarrow$VARIATIONALINFERENCE$(v)$
    **end**
  **end**
  **return** $v_\tau \longleftarrow v'$
**end**

**function** EXPANSION$(v, \mathbf{B})$
  draw randomly an unused action $u'$ on $v$
  for the parent $v$, generate a new child $v'(\mathbf{x}', u')$ with $\mathbf{x}' = \mathbf{B}(u') \cdot \mathbf{x}$
  **return** $v'$
**end**

**function** VARIATIONALINFERENCE($v$)

build the distribution **E** via the probability mass function $\sqrt{\frac{2\ln N(v)}{N(v\prime)}}$

$u' \sim \sigma\left(\kappa_p \ln \mathbf{E}(v') - \gamma \cdot \mathrm{G}_\Delta(v')\right)$

**return** $v'(u')$

**end**

**function** EVAL($v_\tau$, **A**, **B**, **C**, $\delta$)

compute the expected free energy $\mathrm{G}(v_\tau)$ through **A**, **B**, **C**

**return** $\mathrm{G}_\Delta = \delta^\tau \cdot \mathrm{G}(v_\tau)$

**end**

**function** PATHINTEGRATION($v_\tau$, $\mathrm{G}_\Delta$)

**while** $v_\tau$ is not $v$ **do**

$N(v_\tau) \longleftarrow N(v_\tau) + 1$

$\mathrm{G}(v_\tau) \longleftarrow \mathrm{G}(v_\tau) + \dfrac{1}{N(v_\tau)}(\mathrm{G}_\Delta - \mathrm{G}(v_\tau))$

$v_\tau \longleftarrow$ parent of $v_\tau$

**end**

**end**

Taken together, these three terms define the estimated quality of a policy and consider 1) the divergence between preferences encoded in $\mathbf{C}$ and the expected outcomes $\mathbf{A} \cdot \mathbf{x}_\tau$, and expected entropy of observations (respectively, first and second terms of $\mathrm{G}(v_\tau)$), 2) a modulation of the policy quality distribution that controls the stochasticity of action selection, and 3) a confidence bound that regulates exploration.

Note that in standard implementations of active inference, the latter term ($\boldsymbol{E}(v')$) usually encodes prior beliefs about policies that have become habitual (i.e., habitual priors combined with empirical priors furnished by the expected free energy). Instead, in AcT, we use the term ($\boldsymbol{E}(v')$) to promote exploration during tree search, analogous to the exploration bonus used in the UCB1 algorithm for multi-armed bandits [40][72]. Specifically, we define a probabilistic distribution $\boldsymbol{E}(v') = \sqrt{2\, ln\, N(v)/N(v')}$, where $v'$ denotes a child node, $N(v')$ denotes the number of visits of $v'$, and $N(v)$ denotes the number of visits of the parent node $v$. Given this definition of $\boldsymbol{E}$, the probability of every child node decreases if it is visited frequently (i.e., with high $N(v')$) and increases when its number of visits is sufficiently lower than that of the other children (i.e., a large ratio $ln\, N(v)/N(v')$). The effects of $\boldsymbol{E}$ are modulated by an exploration factor $\kappa_p$ and they promote exploration, along with the exploratory (epistemic) term of the expected free energy of policies $\boldsymbol{G}_\pi$ detailed in Section 2. Therefore, analogous to the UCT algorithm [26][70], the Variational Inference stage may select every node with a probability different from zero, increasing in time for less visited states.

### *3.1.2 Second stage: Expansion*

The second (expansion) stage (the function EXPANSION in Algorithm 1) aims to expand the non-terminal leaf node $v_\tau$ selected during the former (variational inference) stage. Expansion of a leaf node $v_\tau$ corresponds to instantiating a new child node $v'$ by implementing a random action $u'$ among those previously unused. Each of these children stands for a future state $\mathbf{x}'$ an agent can visit, according to the transitions defined in the matrix **B**. By adopting a Bayesian terminology, expanding could mean defining new predictable events over the space of future policies, thereby expanding the horizon of possible events. Note that the first (variational inference) and the second (expansion) stages return the same output—a node—but the former stage *selects* a node, whereas the latter *creates* a node. We implemented these two stages into a unique routine, TREEPOLICY in Algorithm 1, analogous to TreePolicy routine [39] in MCTS. In Bayesian statistics, this is not unlike the procedures implicit in nonparametric Bayes, based upon stick-breaking processes that allow for an expansion of latent states [73].

### *3.1.3 Third stage: Evaluation*

The goal of the third (evaluation) stage – implemented by the function EVAL of Algorithm 1 – is to assign a value to the leaf node $v_\tau$ expanded in the previous phase. The evaluation considers the expected free energy $\mathrm{G}(v_\tau)$, a function of the state and the observation associated with $v_\tau$. Note that $\mathrm{G}(v_\tau)$ scores the EFE of the node $v_\tau$, not the sum of the EFEs of all the nodes from the root to the node $v_\tau$. The EFE is then weighted by its 'temporal precision': a discounted factor equal to $\delta^\tau$ that depends on an arbitrary parameter $\delta$ and the depth $\tau$ of the tree node $v_\tau$. The resulting $\mathrm{G}_\Delta = \delta^\tau \cdot \mathrm{G}(v_\tau)$ value, denoted as 'predictive EFE', is finally assigned to $v_\tau$. Please note that, unlike the original MCTS algorithm, evaluating the quality of nodes does not require random policies (or rollouts). This is because the EFE functional computed in the node $v_\tau$ is a sort of an information-based measure that permits simultaneously estimating both the exploitive and the explorative (epistemic) statistical characteristics of the problem configuration represented by the node.

### *3.1.4 Fourth stage: Path Integration*

The fourth (path integration) stage (corresponding to the routine PATHINTEGRATION reported in Algorithm 1) aims to adjust the **G** values of the tree nodes up to the root node $v_t$ by considering the new values obtained during the third (evaluation) stage. The value estimated by EVAL is used to update the quality **G** and the number of visits $N$ of the nodes on the tree path $v_\tau, \cdots, v_t$ obtained during the two phases of variational inference and expansion (which jointly form the TREEPOLICY procedure). Such updating is the statistical analogue of "path-integration" formulations in active inference [10], where one sums up the expected free energy at each time step in the future.

The four stages are repeated iteratively until a criterion is met to provide estimates of the **G** values of a tree node subnet.

### *3.2 Computational resources required by Active Tree Search*

Most on-line planning algorithms employ a belief tree like the one shown in Figure 1, to encode the POMDP problem in a manageable form. The on-line algorithms implement multiple lookahead visits on the tree to plan the next action to execute. A belief tree of depth $D$ contains $\mathcal{O}(|U|^D\ |Z|^D)$ nodes where $|U|$ and $|Z|$ are the cardinalities of the action and observation set, respectively; therefore, the tree size influences the performance of every algorithm relying on belief tree visits in their planning phase. For example, the popular POMCP algorithm [41], which grandfathers the algorithms based on tree visit sampling, is prone to this complexity. The R-DESPOT algorithm [43] generates a subtree of the belief tree of size $\mathcal{O}(|U|^D\ K)$ that encompasses the executions of all policies by $K$ abstract simulations called *scenarios*. In contrast, AcT works on trees with $\mathcal{O}(|U|^D)$ nodes—of considerably less complexity—but uses lossless probabilistic representations for beliefs and observations. This requires greater memory resources and longer updating operations during the run. On the other hand, as noted in [43], POMCP and R-DESPOT, which approximate belief states and outcomes with a particle filter, have problems dealing with large observation spaces because beliefs could collapse into single particles, causing convergence to suboptimal policies.

The total computational cost of AcT depends on the number of simulations controlled through the discount condition $\delta^d < \varepsilon$, and on the EFE computation in the (computationally expensive) Evaluation routine. The number of requisite floating-point operations is proportional to the size of the generative model (e.g., the number of hidden states |S| times the number of outcomes |O|). Note that it would be possible to approximate or amortise EFE computations to reduce computational demands (not explored in this article).

## 4. Results of the simulations using Active Inference Tree Search

We tested the Active Inference Tree Search on three exemplar problems. The first two problems (a deceptive binary tree and a non Lipschitzian function) exemplify deceptive “traps” that are known to challenge UCT and similar algorithms. The notion of “trap” is used in adversarial games strategy search [74] to indicate those states of a game whose instantaneous utility is deceptive, with respect to their future outcomes. The third problem is a POMDP benchmark, the *RockSample* [34], which is often used to evaluate the effectiveness and scalability of planning algorithms as the problem complexity increases[2].

### *4.1 Active Inference Tree Search avoids traps in deceptive binary trees*

Ramanujan et al. [76] noted that the performance of the UCT algorithm is limited in games where the best proximal decisions do not necessarily correspond to winning strategies. This is the case, for example, in chess, where exhaustive search (e.g., minimax) yields better performance than sampling-based approaches. This is due to particular game configurations in which an unfortunate move unavoidably leads to a defeat. These game states are called *traps* because, as noticed above, their instantaneous utility is deceptive with respect to the future outcomes that they lead to. Being able to escape from traps is a crucial feature of successful planning algorithms.

---

[2] Our simulations are carried out using a C++ implementation of AcT that extends a header library described in [75]. The library implements a multi-core parallelisation of the most demanding computational kernels. Note that most of AcT's computational complexity depends on the multidimensional inner products involved in EFE computation and state estimation.

Coquelin & Munos [42] introduced an example challenging problem for sampling-based planning algorithms. This problem has just two actions, $2D + 1$ states parameterised by $D \in \mathbb{N}$, and deterministic rewards. At each time step $d$, one can get a reward of $(D - d)/D$ by choosing action 2; alternatively, one can move forward by choosing action 1. At time $D - 1$, action 1 corresponds to an absorbing state with maximum reward 1 and action 2 to another absorbing state with reward 0. Intuitively, the state space of the problem can be described as a binary tree of depth $D$ (Figure 4). The optimal plan involves always selecting action 1, to move along all the branch levels $d$ and reach the final (maximum) reward. Finding this solution is challenging for sampling methods, as the suboptimal action 2 is much more rewarding in the proximity of the root—and this immediate reward influences planning following the first moves. Coquelin & Munos proved by induction that UCT has a hyper-exponential dependency concerning the depth D of the binary tree. Considering the worst case, it takes $\Omega(\exp(\ldots(\exp(1))\ldots))$ – composed by $D - 1$ exponential functions to get the reward.

To check whether AcT suffers the same limitation, we compared UCT, AcT, and a reduced version of AcT, called FE, which does not use the policy prior beliefs **E** during the exploration stage (or, analogously, with $\kappa_p = 0$). To render this problem suitable for AcT, we reformulated it as an MDP problem, i.e., a Markov Decision Problem that is fully observable. The MDP comprises $2D + 1$ hidden states, a corresponding set of $2D + 1$ observations (consequently **A** is diagonal), action "1" and "2" to move from one state to another (reported in **B**) and a vector **C** where rewards are spread over observations by a probabilistic distribution encoding preferences.

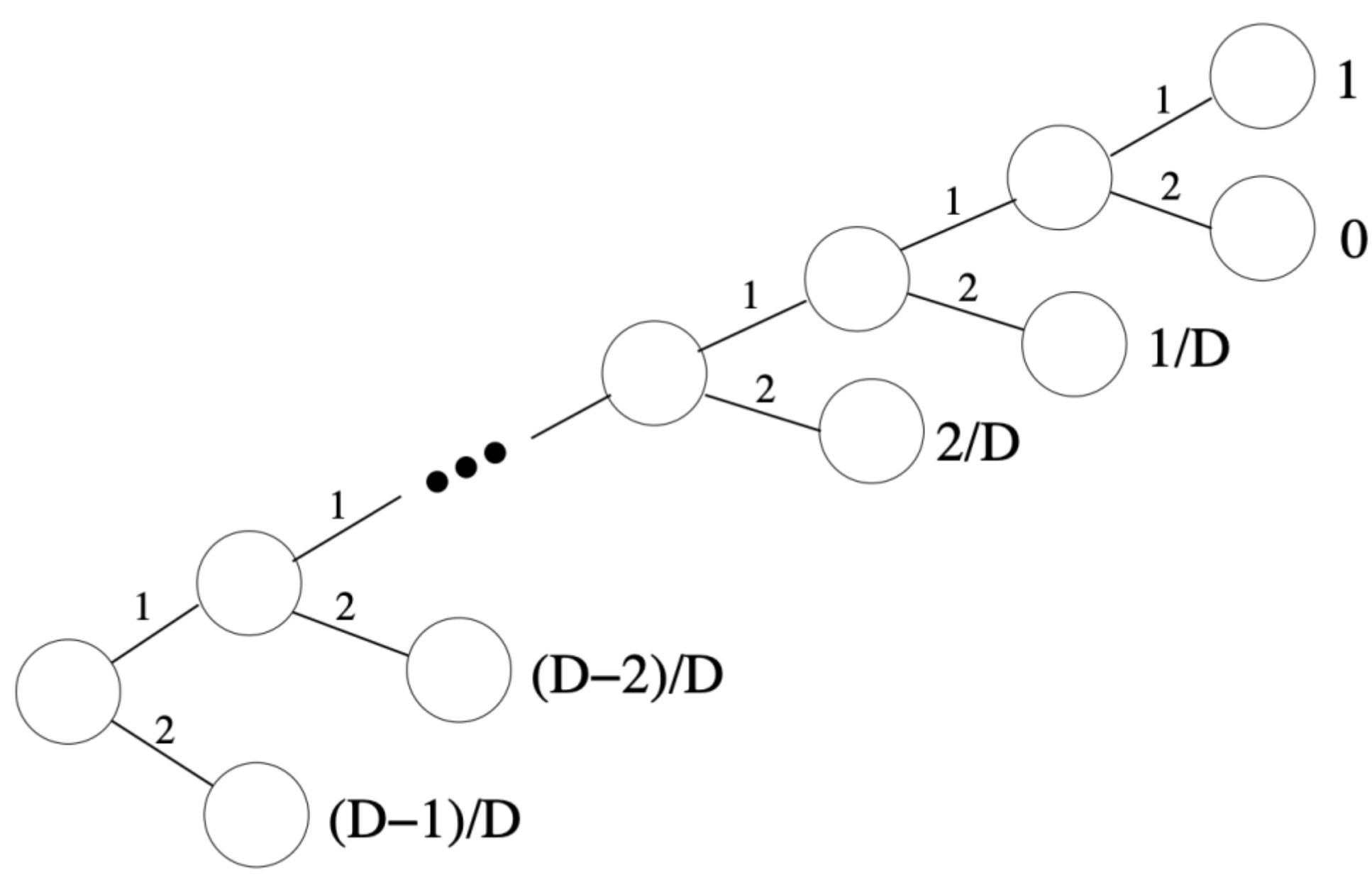

*Figure 4. A binary tree representing the state space of a challenging problem for sampling-based planning methods (adapted from [39]). From the root (left node) toward the deepest level D, action 2 at each level leads to a deceptive leaf node with reward $(D - d)/D$. The optimal policy involves always selecting action 1, which yields reward 1.*

We considered three problems of increasing depth: $D = 10$, $100$, and $1000$. For each problem, we collected the results of 1000 executions of the three algorithms (UCT, AcT and FE) using a fixed number of simulations or playouts (5000). We used a discount factor of $\delta = 0.95$, and set the exploration parameter $\kappa_p = 1$ for both UCT and AcT. The results are shown in Figure 5 by plotting—for each algorithm—the modes of the occupied states, the occupancy probabilities, and the failure rate (defined as the relative difference $(D - d)/D$ between the depth $d$ of the visited state and $D$, with values in $[0,1]$ where 0 states for no failure), as a function of the simulation number.

Our results show that UCT experiences problems starting from $D = 10$ selecting the first deceptive states. Conversely, both algorithms using active inference reached the deepest state, despite performance decreases with greater $D$. FE exhibits a more pronounced greedy behaviour. At the same time, AcT keeps exploring due to the prior distribution **E**. This numerical analysis suggests that AcT has the best performance: compared to the other algorithms—it reaches deeper states of the deceptive tree and does so faster.

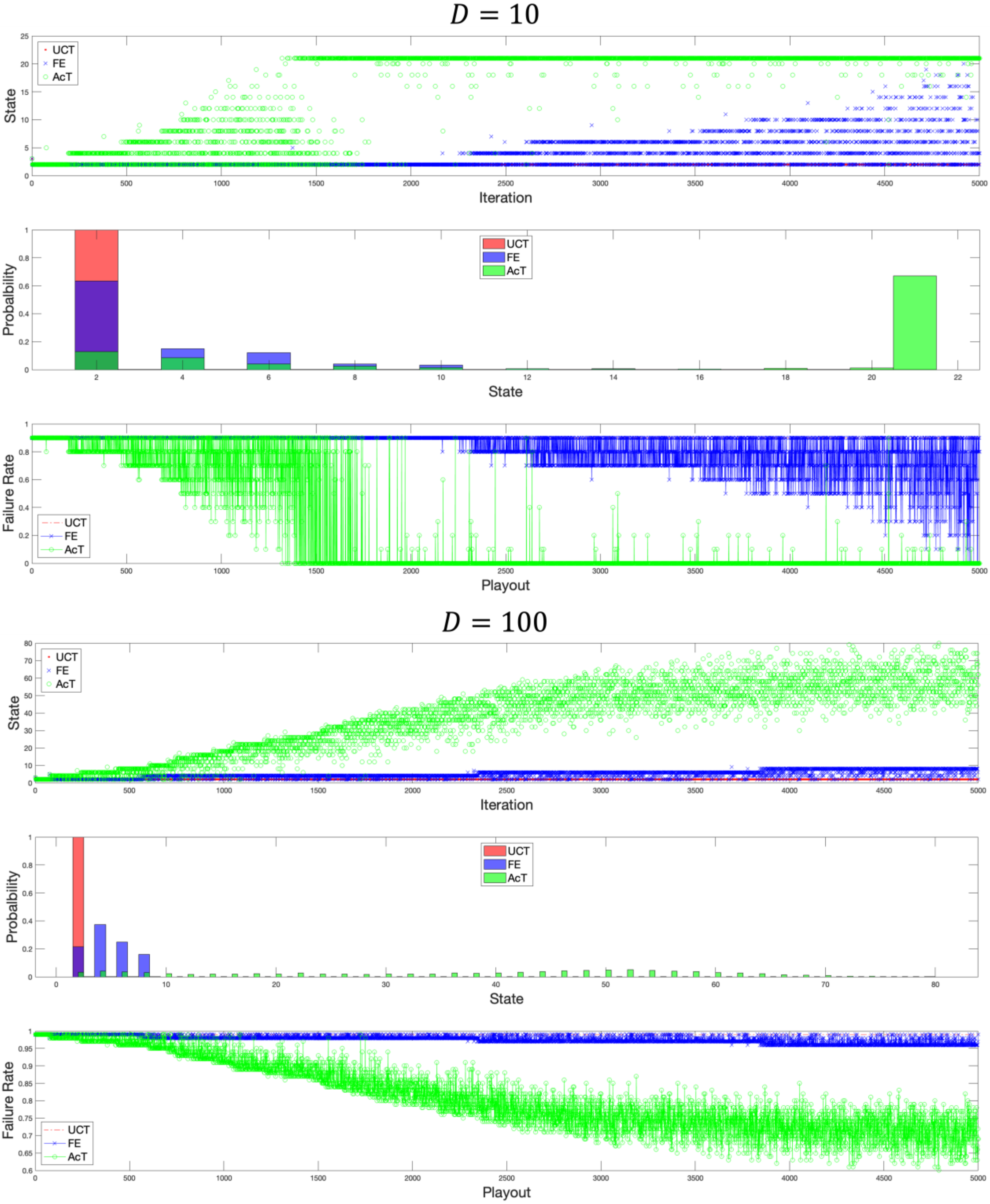

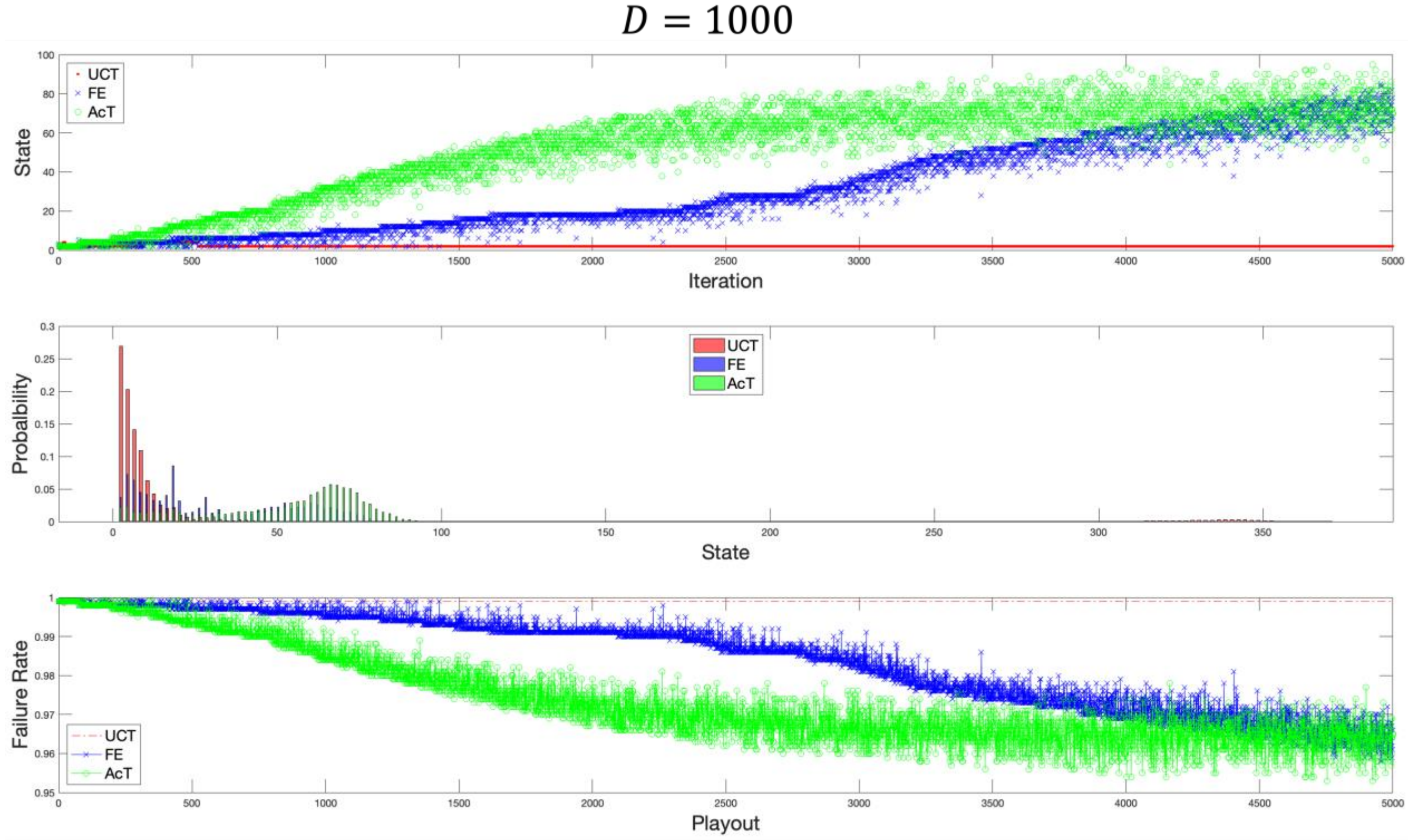

*Figure 5. Experimental results in the deceptive binary tree of [42] of depths $D = 10,\ 100, 1000$ (top, middle, and bottom panels, respectively). Each panel plots state occupation (top), occupation probability (middle), and failure rate (bottom) as a function of the number of simulations (or playouts).*

### *4.2 Active Inference Tree Search reaches an adaptive level of exploration when finding the global maximum of a non Lipschitzian function*

The above problem can be considered illustrative of a whole class of MDP domains on which sampling algorithms manifest shortcomings [77]. These problems are all characterised by the lack of smoothness of the objective or value function, where the notion of "smoothness" corresponds to a well-behaved analytic or continuous value function. Formally, this condition can be expressed through the Lipschitz continuity, according to which a value function $V(s)$ defined over the state-space $S$ is $M$-Lipschitz continuous if $\forall\ s_1, s_2 \in S$, $|V(s_1) - V(s_2)| \leq M\|k(s_1) - k(s_2)\|$, where $M$ is a constant and $k(\cdot)$ is a mapping from $S$ to some normed vector space [78]. The challenge for any optimisation scheme is to find the global maximum of a non-Lipschitzian function. The function:

$$g(x) = \begin{cases} 0.5 + 0.5\left|\sin\dfrac{1}{x^5}\right|, & 0 < x < 0.5 \\ 0.35 + 0.5\left|\sin\dfrac{1}{x^5}\right|, & 0.5 \leq x \leq 1 \end{cases} \tag{10}$$

introduced as a test in [77], has two distinct behaviours over its domain (see panel A in Figure 6). In the (left) interval [0,0.5], there exist numerous global optima, but their functional form is quite rough, whereby in the (right) interval [0.5,1], the function is smooth, but the extrema are suboptimal. In this case, an effective search algorithm should explore every domain region.

As for the binary-tree test used before, we cast this optimisation problem as MDP problem: each state represents some interval $[a, b]$ within this unit square, with the starting state representing [0, 1]. We assume that there are two available actions at each state, the former resulting in a transition to the new state $[a, (b-a)/2\ ]$ and the second resulting in a transition to $[(b-a)/2\ , b]$. For example, at the starting state, the agent has the choice between a "left" action to make a transition to the state [0,0.5] and a "right" action to make a transition to the state [0.5,1]. After it selects the left action, it has a choice between another "left" action to make a transition to [0,0.25] or a "right" action to [0.25,0.5], and so on. Consequently, with increasingly deeper planning trees, the agent explores more fine-grained intervals. An efficient planner should visit the left interval [0,0.5] extensively and deeply (i.e., approach zero), as it encompasses many maxima.

The state-space $S$ can be represented as a binary tree whose depth is constrained by a trade-off set by the condition $b - a < 10^{-5}$. Transitions between states, which move from a state $s$ at a depth $d$ of the binary tree to a state $s'$ at depth $d + 1$ are controlled through the matrix B. Analogous to the function "$g$" shown in Equation 10, also the transition function $s' = B(s, u, t)$ is Lipschitzian but only for domain values larger than 0.5. This is evident by plotting (Figure 6, panel B) the Lipschitzian constant $M$ averaged over all the transitions between two consecutive depths, for $x < 0.5$ and for $x >$ 0.5 (blue and red lines, respectively). For $x \in [0.5,1]$, $M$ increases until it reaches the upper bound value of 10 (for $d > 10$). Instead, for $x \in [0,0.5)$, $M$ shows an exponential behaviour and rapidly reaches much greater values. Each state corresponds to one observation, resulting in an MDP where the matrix $\mathbf{A}$ is diagonal. The a priori distribution $\mathbf{C}$ is computed empirically by considering the value of $g(x)$ in the midpoint of the domain interval encoded by the states. The discount factor $\delta$ was set to 0.95.

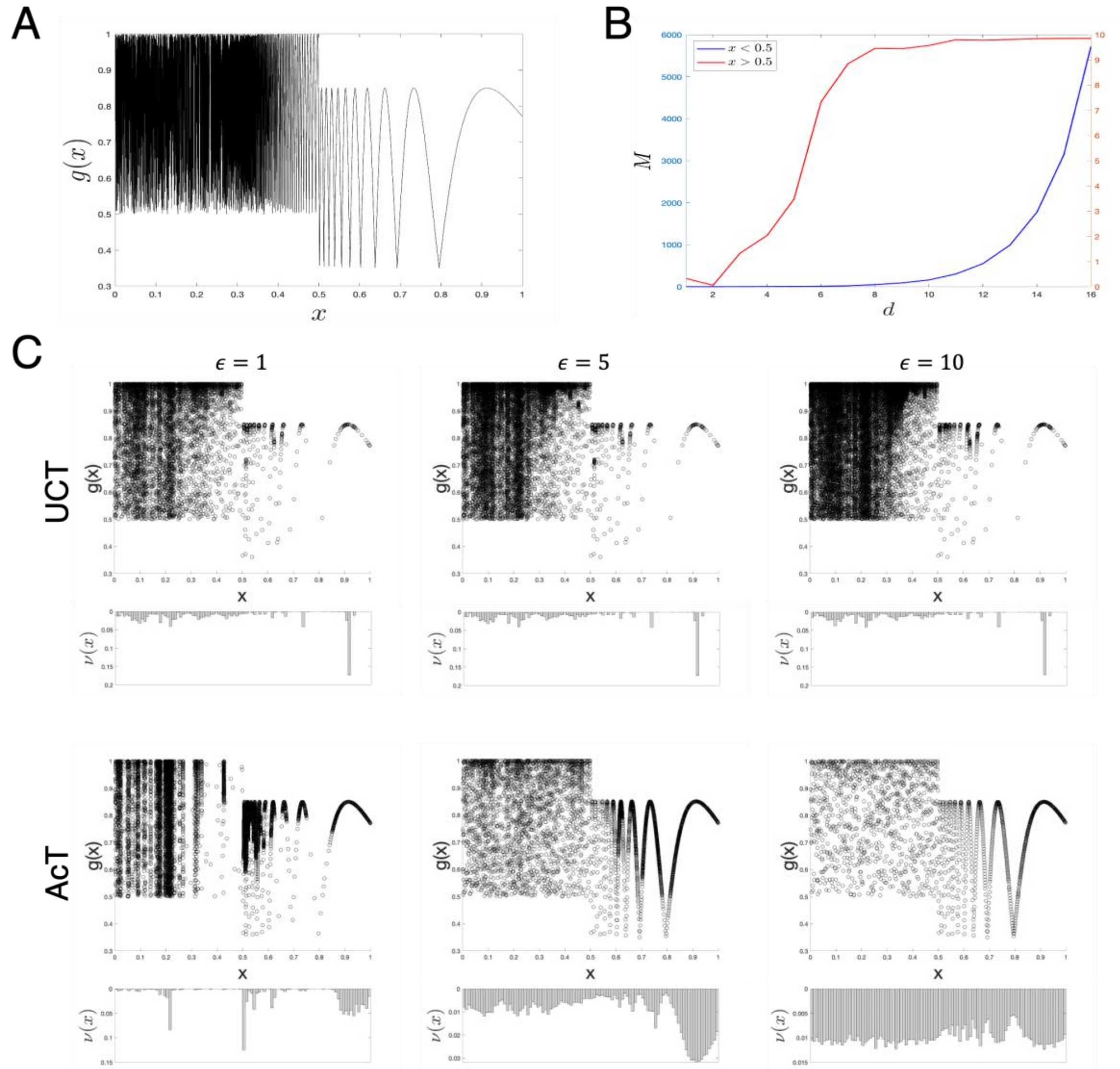

*Figure 6. The function "g" used to test the efficacy of the* AcT *and UCT algorithms in problems with a rough landscape. A) "g" function defined in* $[0,1]$ *is Lipschitz-continuous for values larger than 0.5, yet it is not otherwise. B) The average value of the Lipschitz constant M over all the state transitions at a given depth of the binary tree used to encode the optimisation problem as MDP. The red curve is for* $x \in [0.5,1]$, *whereas the blue plot is for* $x \in [0,0.5]$. *C) A matrix plot whose elements are the scatter plots of the values encoded by the node states visited by UCT and* AcT *(1000 executions each); the bottom parts show the histograms* $\nu(x)$ *of their domain values. The matrix plot arranges the rows by the algorithm (first row for UCT, second row for* AcT*) and columns according to the values of the exploration factor* $\kappa_p$ *set for the executions.*

We compared the UCT and AcT algorithms, for 1000 executions each, with three levels of the exploration factor $\kappa_p$ ($\kappa_p = 1, 5, 10$). Unlike [77], we found that UCT explores the whole domain of $g$, although it mostly visits a state corresponding to an $x$ value around 0.9; see the element (1,1) of the matrix of plots in Figure 6C. AcT explores deeper parts of the tree search (plot (2,1) in Figure 6C) and is able to find maxima in the whole domain of $g$. Compared to UCT, AcT shows greater exploitative behaviour in correspondence with specific significant $x$ values, for instance, 0.5, 0.9, and 0.2 (that are the modes of the visited $x$ distributions).

This optimisation problem illustrates the effects of the exploration factor $\kappa_p$ on algorithm performance. Reading Figure 6C out by columns, one can evaluate the effects of $\kappa_p$ on the algorithms UCT (first row) and AcT (second row). The performance of UCT remains relatively stable across all the values of $\epsilon$, in the sense that despite the increase of exploration with greater values of $\kappa_p$, the statistical distribution $\nu(x)$ of the visited domain points remains unchanged. Instead, the effects of the parameter $\epsilon$ on AcT are more significant. For $\kappa_p = 1$, AcT shows limited exploration, similar to the FE algorithm used in the deceptive binary tree example. This is because, in this particular problem, the term used to sample the actions (related to the **E** and controlled by $\kappa_p$) is numerically much smaller than the one related to the policy value **G**. When $\epsilon$ is set to 5 or 10, AcT explores significantly more and (with $\kappa_p = 10$) it visits the $g$ codomain uniformly. In this latter case, AcT also visits the unrewarded branches of the binary tree, even if this implies a reduction of performance; this becomes apparent by noticing that $\nu(x)$ is almost flat for $\kappa_p = 10$.

### *4.3 Active Tree Search in large POMDP problems: the case of RockSample*

*RockSample*$(n, k)$ [34] is a well-known benchmark problem for assessing POMDP solvers and their scalability. It simulates a rover whose task is collecting *samples* of $k$ scientifically valuable rocks deployed on an $n \times n$ alien soil grid—and then leaving the area. Samples come in two varieties: valuable or invaluable. The rover earns a reward for each valuable sample it collects and a penalty for each invaluable sample. The rover knows the locations of the rocks but can only evaluate whether they are valuable via a long-range sensor, whose measures are affected by an error, which increases exponentially with the distance between the rover and the rock examined. We considered two variants of this problem, with $(n, k)$ equal to $(7,8)$ and to $(11,11)$.

Representing this problem in a format suitable for active inference is straightforward. In principle, the problem state space $S$ can be factorised to reduce the total number of states [79]. Still, we decided to retain both variants without factorised representations to leverage the problem's difficulty. Therefore, the cardinality of the state space is $|S| = n^2 * 2^k + 1$ (12,544 in the case (7,8) and 247,808 in (11,11)), necessary to encode every possible combination of locations and the scientific value of the rocks plus an additional "exit" state. This cardinality is needed to define the initial belief state **D** and the transition state matrix **B**, which is also conditioned on the control state (actions) $a \in U$ that the agent can make.

The set $U$ contains the four actions (go north, go south, go east, go west) that the agent uses to move around the square, $k$ actions that the agent uses to evaluate the rocks remotely (one action for each rock), and a sampling action to collect a rock sample. Observations are factorised into three factors, which relate to the positions on the grid, the configuration ($2^k$) encoding the scientific quality of the rocks, and their associated rewards, respectively. Accordingly, the likelihood **A** is decomposed into three factors, each one encoded as a cubic matrix (generally as a tensor when the state space, in turn, is subdivided into factors), where the first dimension represents the observations, the second the states, and the last the actions. Introducing a dependency of **A** on the actions is uncommon in active inference but useful in many POMDP problems, including *RockSample*$(n, k)$. This is because the observation one gets by sampling a rock (with an action $k$) is a function of the distance from the rock; encoding this contingency would require a considerable number of states if one does not express **A** as a function of actions. Reward contingencies expressed in **A** are action-dependent, too, as the agent obtains a "good" observation when it samples a good rock (and a "bad" observation otherwise)—and when it exits the game. Finally, **C** encodes preferred observations and comprises three modalities: in the first two, observations are uniformly preferred, while in the last, they are drawn from a Bernoulli distribution, with a success probability almost equal to 1. See the Appendix for an example generative model for *RockSample*, with $n = 2$ and $k = 1$.

We used the same parameters for both *RockSample*$(7,8)$ and *RockSample*$(11,11)$. We used a discount factor $\delta$ equal to 0.95 and a 'discount horizon' $\varepsilon$ of 0.4 so that the depth $d$ of the tree search developed during planning is about 19 steps (a threshold computed by considering that $\delta^d < \varepsilon$). We evaluated 1000 executions of AcT, with different seeds from a pseudorandom number generator and with different (random) arrangements of rocks in the grid.

In keeping with previous works [41][43], we augmented the AcT algorithm with a domain-specific, heuristic policy that prioritises some selected actions during the simulations. Specifically, the heuristic policy prioritises actions that approach the rocks with more "good" observations and actions that check rocks with uncertain outcomes (ensuing from inconsistent observations). When the rover is in the same place as a rock evaluated as "good" by most observations, the heuristic policy prioritises sampling actions. Finally, when all the rocks in the scenario have been sampled, the heuristic policy prioritises actions heading toward the exit.

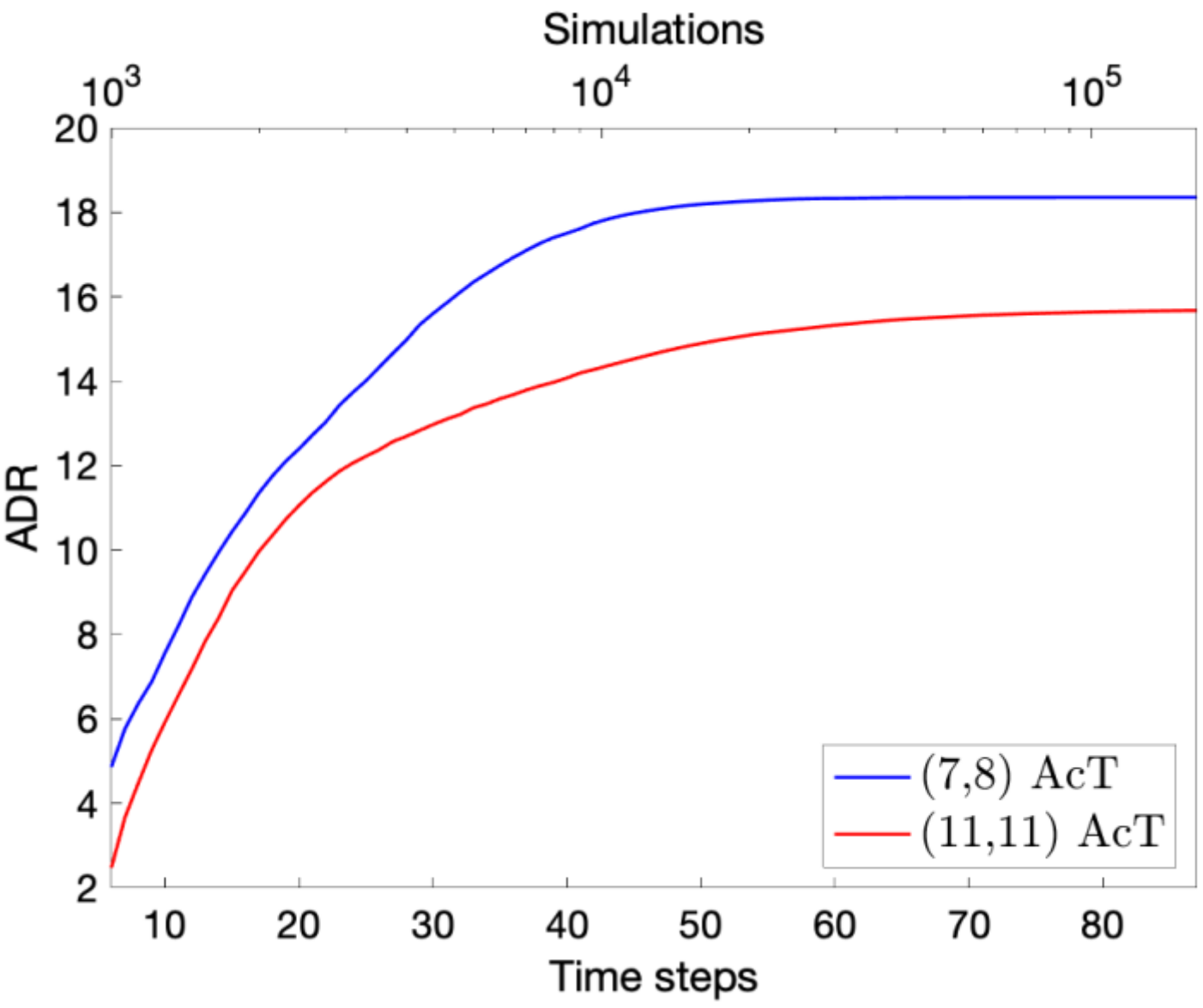

*Figure 7. Total discounted reward achieved by the* AcT *algorithm augmented with the heuristic policy for RockSample*$(7,8)$ *and RockSample*$(11,11)$ *(in blue and red, respectively). Results are shown as a function of the time steps (at the bottom axis) and the number of simulations (in logarithmic scale, top axis). All results are averaged over 1000 executions.*

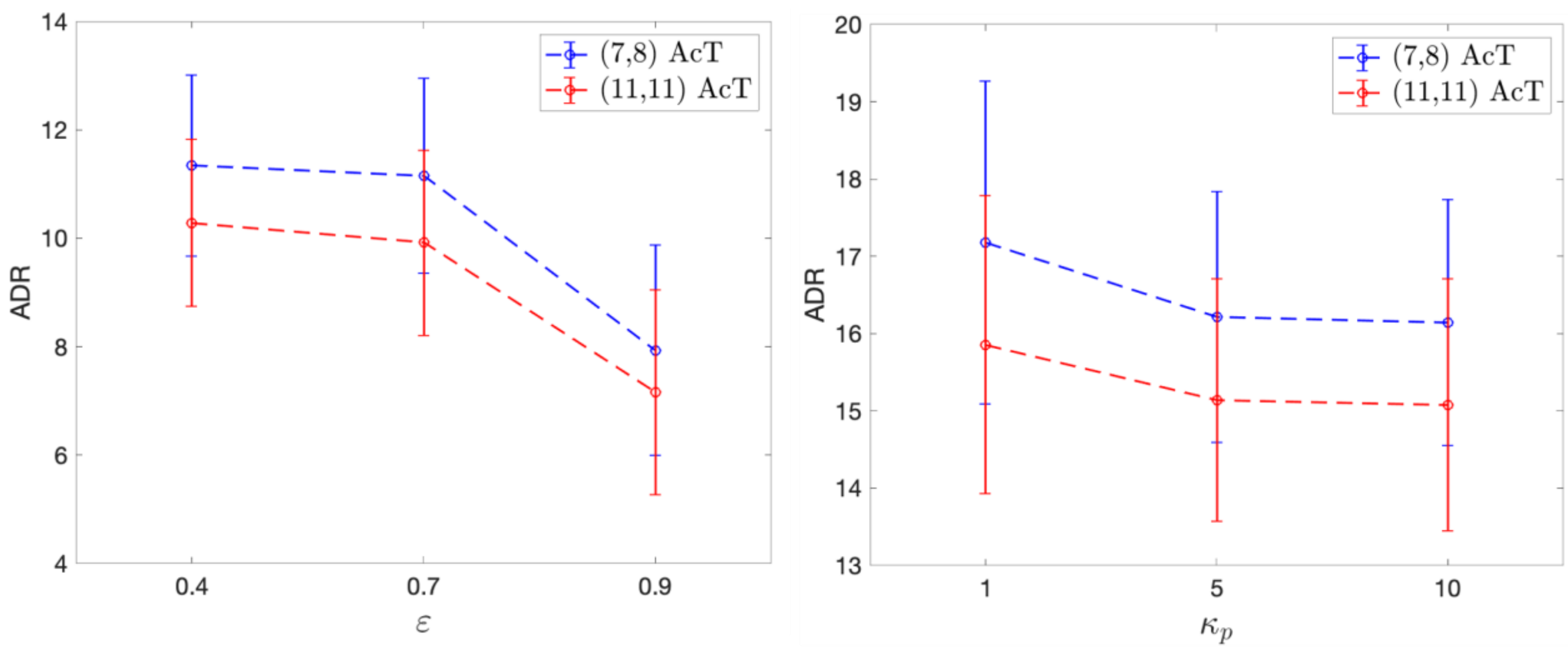

*Figure 8. Total discounted reward (ADR) averaged over 1000 execution, obtained by* AcT *as a function of its control parameters. Left panel: ADR as a function of the discount horizon* $\varepsilon$ *(with* $\kappa_p = 1$*). Right panel: ADR as a function of the exploration factor* $\kappa_p$ *(with* $\varepsilon = 0.4$*). In both panels, the blue and the red lines represent the results obtained for* $RockSample(7{,}8)$ *and* $RockSample(11, 11)$*, respectively.*

Figure 7 shows the model performance, expressed as the total discounted reward (ADR) $\bar{R}_t^{\delta} = \frac{1}{N}\sum_{i=1}^{N}\sum_{\tau=0}^{t}\delta^{\tau} r_i\,(\tau)$ averaged over $N$ executions, as a function of time steps $t$ required to complete the task. In $RockSample(7{,}8)$, AcT achieves $18.35 \pm 4.17$ ADR in 38.92 time steps and 52,672.5 simulations (on average). In $RockSample(11{,}11)$, AcT achieves $15.71 \pm 3.86$ ADR in 74.93 timesteps, with 238,461 simulations (on average)[3]. In both $RockSample(7{,}8)$ and $RockSample(11{,}11)$, AcT scales up smoothly with the size of the problem, as evidenced by the fact that the algorithm's performance shows the same trend in both problems.

We analysed the performance of AcT in $RockSample(7{,}8)$ and $RockSample(11{,}11)$ as a function of the horizon discount parameter $\varepsilon$ (Figure 8A). As expected, the performance of AcT decreases when $\varepsilon$ increases (and consequently, the maximum depth of the planning trees decreases). It could be noted that there is a threshold (which is plausibly domain-specific), after which the increase of $\varepsilon$ becomes catastrophic. Indeed, AcT preserves its effectiveness between $\varepsilon = 0.4$ and $\varepsilon = 0.7$ but becomes unsuccessful with $\varepsilon = 0.9$ when the planning tree becomes excessively small.

[3] The AcT algorithm without the domain-specific heuristic policy achieves significantly lower scores: $13.4251 \pm 5.89$ ADR for $RockSample(7{,}8)$ with $\varepsilon = 0.7$, and $6.15488 \pm 3.82329$ ADR for $RockSample(11{,}11)$ with $\varepsilon = 0.9$. Please note that these results are obtained with $\varepsilon$ values significantly higher than those reported in the main simulation ($\varepsilon = 0.4$). This is because using $\varepsilon = 0.4$ without the heuristic policy entails significant memory demands. Increasing $\varepsilon$ permits decreasing memory demands by constraining the depth of the planning tree.

Furthermore, we analysed the performance of AcT in *RockSample*$(7,8)$ and *RockSample*$(11,11)$ as a function of the exploration factor $\kappa_p$ (Figure 8B) while keeping the horizon discount parameter $\varepsilon$ fixed ($\varepsilon = 0.4$). The performance of AcT decreases when the exploration factor $\kappa_p$ is increased from 1 to 5 or 10 (the results for $\kappa_p = 5$ and $\kappa_p = 10$ are almost indistinguishable). This result indicates that an explorative approach is ineffective in *RockSample*, as it leads AcT to overextend the tree width, disregarding the most rewarding branches.

In sum, the simulations reported in this section provide a proof of principle that active inference can be scaled to deep planning problems. In the next section, we consider the Active Inference Tree Search algorithm from the perspective of neuronal dynamics.

### *4.4 Simulated neuronal dynamics of Active Inference Tree Search*

This section illustrates the usage of Active Inference Tree Search to simulate behavioural and neurophysiological responses during human planning. To exemplify this, we apply Active Inference Tree Search to "Tiger": a popular POMDP problem introduced in [80], to illustrate the importance of epistemic, information-gathering actions (that aim to acquire information to reduce uncertainty) during planning. In the Tiger problem, an agent stands in front of two doors and has to decide which one to open. The agent knows that one of the two doors hides a treasure, whereas the other conceals a tiger. If the agent opens the door with the treasure, it receives a reward, but if it opens the door with the tiger, it receives a penalty. The agent does not know where the tiger is but can resolve uncertainty by "listening for animal's noises" (which induces a small cost).

The domain of this problem is usually represented as a POMDP with 2 states (tiger behind the left or right doors), 3 actions (to open the two doors or listen), and 2 observations (reward or penalty). To ensure compatibility with previous active inference studies, we recast the problem as a T-maze with 8 states, 4 actions and 16 observations [10]. The 8 states result from the multiplication of 4 locations times 2 hidden contexts. The 4 locations correspond to the centre (i.e., start location), the left and the right arms (analogous to the two doors, with treasure and tiger, respectively), and the lower arm (analogous to a listening location, where a cue can be found that discriminates the tiger location). The 2 hidden contexts correspond to the 2 possible reward locations (i.e., the reward at the left or the right arm, respectively). The 4 actions move the agent deterministically to each of the 4 locations (but cannot change hidden context). Finally, the 16 observations result from the multiplication of 4 positional observations (that correspond 1-to-1 to the 4 locations) by 4 outcomes (i.e., reward, penalty, cue for the tiger at left, and cue for the tiger at right) that are obtained in different states, see below.

The matrices **A**, **B**, and the vectors $\mathbf{C}$ and $\mathbf{D}$ specify the agent's generative model. The (likelihood) matrix $\mathbf{A}$ is a probabilistic mapping from states to outcomes. It specifies that the centre location provides an ambiguous cue (i.e., a cue that is identical if the agent is in either of the 2 hidden contexts and hence does not provide any information about the reward location). Furthermore, it specifies that the lower arm provides a disambiguating cue—that discloses which of the 2 hidden contexts the agent is in (and hence the reward location). Finally, the likelihood specifies that if the agent is in the first hidden context ("reward at the right arm"), the right and the left arms provide a reward and a penalty, respectively, with probability $p = 0.90$. On the contrary, if the agent is in the second hidden context ("reward at the left"), the right and the left arms provide a penalty and a reward, respectively, with probability $p = 0.90$.

The $\mathbf{B}(u)$ (transition) matrices define 4 action-specific stochastic transitions between states. These move the agent deterministically to each of the 4 locations (but cannot change its hidden context). However, there is a peculiarity: given that the task ends when the agent is in one of the upper arms (i.e., opens one of the two doors), we consider the corresponding hidden states as absorbing states that cannot be left, whatever the action.

The vector $\mathbf{C}$ encodes the probability mass over preferred outcomes. It is determined by applying the Softmax function over a utility vector having 2 and $-2$, respectively, for rewarding and penalty outcomes and zeros otherwise. Finally, $\mathbf{D}$ represents the agent's belief about its initial state. The agent knows that it starts from the centre location, but—crucially—it does not know in which of the 2 hidden contexts it is in (i.e., it does not know the reward location, left or right). This is why it is optimal for the agent to go to the lower arm to solicit a cue (i.e., "listen") that disambiguates the hidden context before deciding whether to visit the left or right arms. As in the previous simulations $\delta = 0.9$.

Figure 9 illustrates the results of Active Inference Tree Search simulations, in which the reward is in the right arm. The upper panel shows the agent's state-belief distribution over time and the true states (cyan circles). At the first epoch, the agent knows it is in the centre location but does not know its current context. This is evident when considering that in the first column of the upper panel, the belief distribution spans states 1 and 2 (i.e., centre location in the first and second hidden contexts). The agent then selects an action to visit the lower arm to collect a cue; and it discovers that the hidden context is the first (reward at the right arm). This is evident in the second column of the upper panel, where the belief distribution is concentrated in state 7 (i.e., lower arm, first hidden context). Note that the agent decided to explore the lower arm to secure a cue instead of guessing which of the two arms is rewarding. Although this entails a cost (a delay in reaching the reward location later), this "epistemic behaviour" ensures the selection of the rewarding arm at the next epoch, see the third column of the upper panel. This epistemic behaviour emerges automatically in active inference (K. J. Friston, 2010) because policy selection considers the expected reduction of uncertainty along with utility maximisation (see Equation (7) of the section "active inference").

The second panel of Figure 9 shows the "search trees" that AcT generates during each epoch. The left picture of the second panel shows the search tree generated during the first epoch, where the thickness of the edges connecting levels reports the probabilities of going to one of the 4 locations. The preferred plan at depth one is to make an "epistemic" move to visit the lower arm. At depth two, the two preferred actions are to visit the lower arm and (to a lesser extent) the centre location. This is because the tree search has not yet received any observation from the generative process and, therefore, has no information about the tiger location—hence, it avoids states that include potential penalties.

The centre graphic of the second panel shows the search tree generated during the second epoch after the agent has visited the lower arm and has observed an informative cue. At this point, the agent constructs a new search tree where the plan to reach the right arm is highly probable. The choice remains unchanged at the last epoch (see the right picture of the third panel), and the agent collects the rewarding outcome.

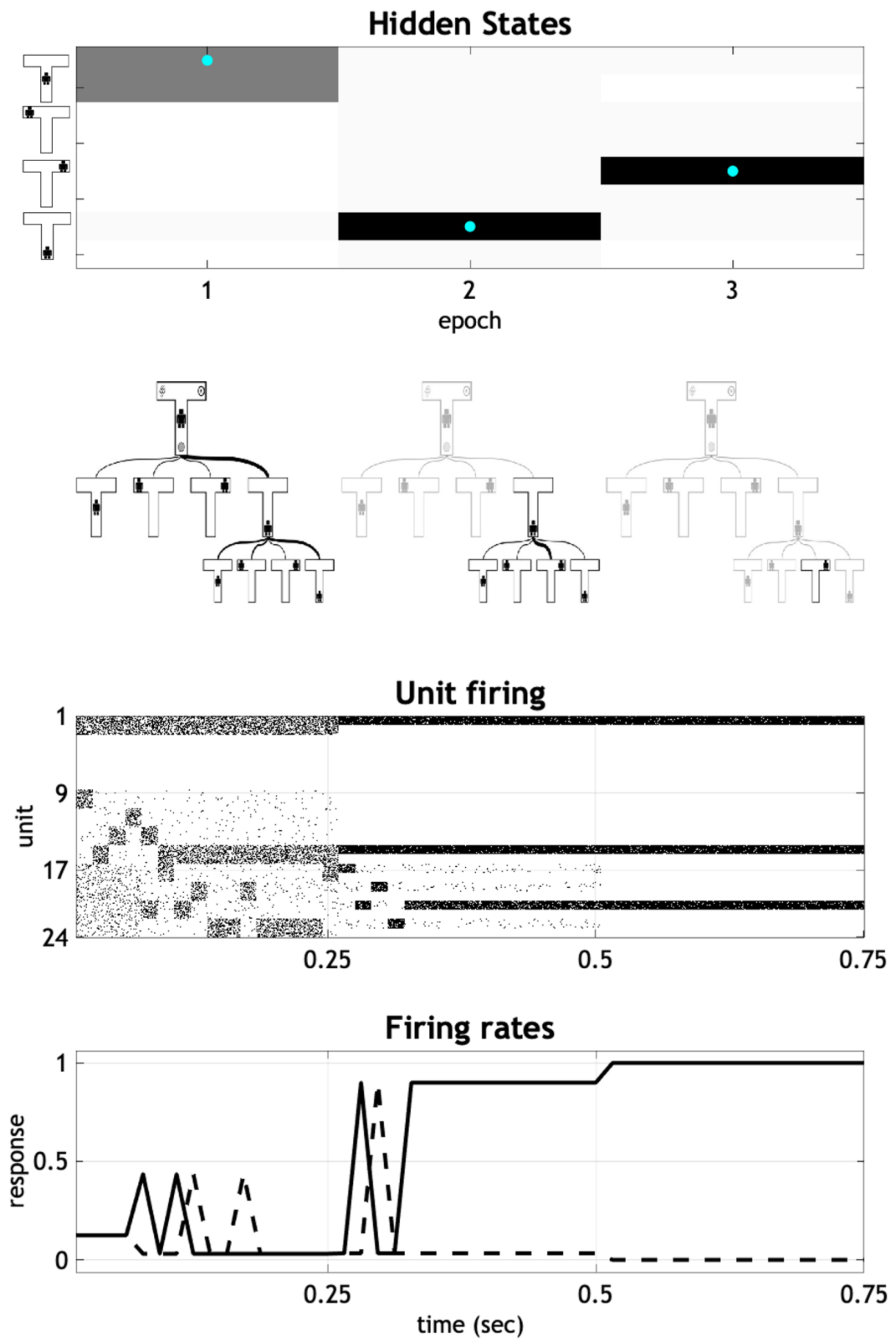

*Figure 9. Simulated behavioural and neuronal responses of Active Inference Tree Search in the "Tiger" problem, when the tiger is behind the left door. The first (top) panel shows the belief distribution over the eight hidden states (4 locations, each one in two possible contexts: "reward at the right" on top and "reward at the left" on bottom) as a function of time steps or epochs. The*

*second panel illustrates the search trees generated by AcT in the three epochs of the simulation. The cyan dots represent the actual states the agent is in, at each time step. The third panel shows simulated neurophysiological responses associated with planning, following the encoding of probabilistic states by neuronal activity described in [11]. This assumes that, at each epoch, an agent represents probabilistic beliefs about the present, the past and the future, in distinct neuronal populations. These are displayed as firing rates of 24 (sets of) single cell units (neurons) that encode hidden states over the 3 epochs in a raster plot format. Specifically, (sets of) units 1 to 8 are related to hidden states of the first epoch, (sets of) units 9 to 16 to hidden states of the second epoch, and (sets of) units 17 to 24 to hidden states of the third epoch. The first two (sets of) units for each epoch encode the central location, i.e., (sets of) units 1 and 2 encode the central location in the first epoch, (sets of) units 9 and 10 encode the central location in the second epoch, and so on… Finally, odd and even units encode the "reward at the right" and "reward at the left" hidden context, respectively. For example, (sets of) units 1 and 2 encode the central location in the first epoch, when the context is "reward at the right" and "reward at the left", respectively. The fourth panel plots the firing rates of two neurons encoding the right arm (solid line) and left arm (dashed line), on the third epoch. These are the states that will be finally selected (right arm) and unselected (left arm). These correspond to neurons at lines 21 and 19 in the third panel, respectively. Please see the main text for an explanation.*

The third panel of Figure 9 illustrates simulated neurophysiological responses during the simulation shown in the first two panels. We assume that outcomes are sampled every 250 ms: a timescale compatible with hippocampal theta cycles, where place cells representing current and prospective locations can be decoded [81–83]. The figure illustrates a raster plot of simulated neuronal activity for units encoding hidden states. The image is organised as a matrix, with 24 rows/neurons (4 locations times 2 contexts times 3 epochs, corresponding to the planning horizon) and as many columns as the number of rollouts—as implemented by AcT during the three decision epochs (16 in this simulation). In other words, the presence of 24 rows/neurons indicates that at each epoch, the agent represents its current epoch and the two closest epochs (e.g., during the first epoch, it also represents the subsequent two epochs; during the second epoch, it represents the previous and the subsequent epochs). Furthermore, separate neuronal populations encode the same hidden states at different epochs (e.g., the first hidden state at the first and second epochs correspond to the 1st and the 9th rows/neurons, respectively). In effect, this endows the agent with a form of working memory that is both predictive and postdictive.

The rows and the columns of the third panel can be grouped to cluster the matrix in $3 \times 3$ blocks of length 8 and 16, respectively. In this format, the elements shown in the main diagonal of the block matrix are beliefs about the present and correspond to the hidden states shown in the first panel. The elements shown in the upper and lower diagonal blocks correspond to (postdictive and predictive) beliefs about the past and the future, respectively. Note that the elements under the main diagonal correspond to the beliefs shown in the search trees of the second panel.

The fourth panel of Figure 9 reports the simulated firing rates of two selected units, which correspond to the states representing the left (dashed line) and the right arm (solid line) during the third epoch. These are the states that will be visited (right arm) and not visited (left arm) during the third epoch.

Initially (first column of the fourth panel), both units have the same firing rates because the agent is uncertain about the state it will visit next. However, this uncertainty is resolved during the second epoch and confirmed during the third (second and third columns, respectively). It is evident from this panel that expectations about future visitations (corresponding to the firing rates of the two units) diverge during the epochs, following a stepwise evidence accumulation [11].

These simulations exemplify the possibility of establishing a mapping between algorithmic methods of AcT and neuronal processes relevant to neuroscience. For example, the neurophysiological responses shown in the last two panels of Figure 9 exemplify prospective (and retrospective) representations of states that have been consistently reported in rodents [84–87], monkeys [88–90], and humans [91,92] engaged in a sequential decision or navigation tasks. From a cognitive science perspective, postdictive and predictive representations of this sort could be construed as working memory for past or future events or goals [93].

While computational modelling is widely used in neuroscience, there is still a paucity of methods that can both address large-scale problems relevant to AI and generate predictions relevant to neuroscience. Some recent studies using powerful deep learning [94–96] and Bayesian methods [97] are already bridging this gap, but they mostly address specific domains, such as visual perception and motor control. Addressing tasks such as *RockSample* or Tiger requires designing complete agent architectures instead, as exemplified by AcT (this paper) and deep reinforcement learning models [98]. AcT and deep reinforcement learning models appeal to different principles—e.g., free energy minimisation versus reward maximisation; inference versus trial-and-error learning; appeal to the Bellman optimality principle versus variational principles of least action, and so on—to design agent architectures that solve complex tasks, hence speaking to different views of neuronal dynamics. Comparing the empirical validity of these assumptions side-by-side is an important objective for future research.

## 5. Discussion

Model-based planning is a widely interdisciplinary topic. However, synthesising ideas and methods from disciplines as diverse as AI, machine learning, and cognitive and computational neuroscience has been challenging, given their different focus (e.g., scalability and efficiency in AI, biological realism in neuroscience).

Here, we offer a significant step in this direction by extending a prominent neurobiological theory of model-based control and planning—active inference—to scale it up to POMDP problems of much larger size. This extension exploits tree search to elude the extensive evaluation of action policies often countenanced in active inference. The theoretical synthesis of active inference and tree search planning methods—called *Active Inference Tree Search*—benefits both. On the one hand, augmenting active inference with tree search methods permits the realising of a novel and appealing process model for approximate planning. This renders it scalable and potentially useful for explaining bounded forms of cognition and reasoning [56,99–101]. On the other hand, active inference provides a theoretically motivated and biologically grounded framework to balance exploration and exploitation, which contextualises heuristic methods widely adopted in tree search planning, permits avoiding rollouts (as in Monte-Carlo methods) and obtains remarkable results in challenging POMDP problems.

We validated *Active Inference Tree Search* in three simulative studies. The first study's results show that AcT successfully addresses deceptive binary trees that challenge most sampling-based planning methods, as it requires an accurate balance of exploration and exploitation. In AcT, the balance of exploration and exploitation depends implicitly on a single free energy functional used for policy evaluation. The results of the second study confirm the adaptivity of the exploration strategy used in AcT. They suggest that AcT can resolve challenging problems whose value functions are not smooth and are, therefore, challenging to explore systematically. The results of the third study show that AcT can successfully address POMDP problems (here, *RockSample*). The performance of AcT scales gracefully with problem size and is quite comparable with state-of-the-art POMDP algorithms, such as SARSOP [32], AEMS2 [31], POMCP, and DESPOT [43]; see also [41,102].

Finally, we used *Active Inference Tree Search* to simulate neuronal responses during a representative planning task. This simulation illustrates the ability to map the algorithmic-level planning dynamics of AcT to neuronal-level representations putatively found in the hippocampus (and other areas, such as the prefrontal cortex) of animals that solve equivalent tasks [81–83]. Indeed, active inference originated in computational and systems neuroscience, intending to characterise brain processes from a normative perspective. AcT retains the neurobiological motivation of active inference while aiming to expand it to large-scale problems that previous implementations could not address. Using AcT to address large scale planning problems and explain neuronal activity can help establish a much-needed bridge between AI and computational neuroscience.

In sum, our results show that AcT can deal with large state spaces, despite using off-the-shelf active inference methods that were not optimized for such problems. However, the performance of bespoke, state-of-the-art POMDP algorithms [41,102], such as DESPOT [43] remains superior. An important challenge — for future work – is adopting AcT to address vision-based POMDP benchmarks, which deal with high dimensional image input. Addressing vision-based POMDP tasks might be feasible by adapting solutions from previous studies demonstrating active inference "from pixels" [103–105] and the possibility of using target images (or imagined images) as preferred outcomes for robotics [106,107]. Another objective for future research would be benchmarking AcT by comparing it with state-of-the-art POMDP schemes, especially those using deep learning. In this respect, a promising direction consists not just of comparing AcT with POMDP and deep learning methods but also hybridizing them, given that they might have complementary strengths. AcT and state-of-the-art POMDP algorithms—based on reinforcement learning— rest on different principles (variational principles of least action and Bellman optimality, respectively) and use different objectives; namely, expected free energy and expected reward. However, there are various aspects of AcT that might be potentially useful for planning in large scale domains. First, the minimisation of free energy naturally encompasses both exploration and exploitation under a single imperative, alleviating the need for explicit exploration terms—or the need to balance expected information gain against expected reward. The inherent exploration capabilities of AcT are illustrated in the better performance, compared to UCT, when addressing deceptive binary trees, as in Figure 4 (noting that UCT could achieve the same results with a higher exploration constant). Intrinsic exploration capabilities could be especially beneficial in tasks that are ambiguous or lack explicit rewards [67,108]. Second, AcT—along with other methods to formulate reinforcement learning and control problems as probabilistic inference [109], such as planning-as-inference [15] and KL control [18]—uses a probabilistic formulation that could be leveraged in Tree Searches. For example, the expected free energy scores the probability of alternative policies and can therefore be used to automatically terminate searches at a particular node in the tree, in a way that eschews sampling. Third, representing preferences (or rewards) as priors over observations provides flexibility in modeling reward functions; especially in complex nonstationary environments [108,110]. One aspect of casting rewards as prior preferences is that rewards can be specified as constraints over all outcomes, enabling a form of multiple constraint satisfaction.

On the other hand, AcT extensions could benefit from incorporating tools from deep learning, such as methods to approximate or amortise EFE computations to reduce computational demands. Furthermore, future AcT extensions could potentially address continuous spaces by adapting solutions developed for Monte Carlo Tree Search, such as the UCT extension called 'Progressive Widening' [111], the use of a kernel regression for estimating the utility of a random continuous action [112], or the value-gradient based variant proposed in [113]. These solutions require modifications to the tree search procedure, the nature of the distributions involved, and their evaluation. It remains to be seen if these methods can be deployed to improve AcT, and whether they could be integrated (or augmented) with active inference to handle mixed (i.e., discrete and continuous state space) generative models [8,114,115]. These extensions of AcT might shed light on the conceptual and mathematical relations between different formulations of the same planning tasks—as reinforcement learning, control or inference problems [13]—hence, creating synergies between distinct research fields.

**Declaration of conflict-of-interest**

The authors have no conflict-of-interest to declare.

**Acknowledgements**

KJF is supported by funding for the Wellcome Centre for Human Neuroimaging (Ref: 205103/Z/16/Z) and a Canada-UK Artificial Intelligence Initiative (Ref: ES/T01279X/1). KJF and GP are supported by the European Union's Horizon 2020 Framework Programme for Research and Innovation under the Specific Grant Agreement No. 945539 (Human Brain Project SGA3). GP is supported by the European Research Council under the Grant Agreement No. 820213 (ThinkAhead), the European Union's Horizon 2020 Framework Programme for Research and Innovation under the Specific Grant Agreement No. 952215 (TAILOR), the Italian National Recovery and Resilience Plan (NRRP), M4C2, funded by the European Union – NextGenerationEU (Project IR0000011, CUP B51E22000150006, "EBRAINS-Italy"; Project PE0000013, "FAIR"; Project PE0000006, "MNESYS"), and the Ministry of University and Research, PRIN PNRR P20224FESY. The GEFORCE Quadro RTX6000 and Titan GPU cards used for this research were donated by the NVIDIA Corporation. The funders had no role in study design, data collection and analysis, decision to publish, or preparation of the manuscript.

## Appendix

### A.1 An example generative model: the case of *RockSample*(2,1).

Here, we provide an example of how a POMDP problem can be represented in Active Inference. We focus on the $RockSample(n, k)$ problem with $n = 2$ and $k = 1$. In this scenario, the rover explores an alien soil grid of side 2, with a unique rock to analyse, see Figure A.1. The rover starts its exploration from the top-left corner and has to reach the exit on the right side after (optionally) collecting a sample of the rock. The position of the rock in the maze and its quality ("good" or "bad") are selected randomly. In our example, the rock is placed in the bottom-right corner of the maze, and its quality is "good". Figure A.1 shows the initial configuration of the problem: the locations of the rover are denoted with "$R_i$" (with $R_1$ being the initial location), the location of the rock is denoted with "€", the exit state is denoted with "EXIT", and the grid border (that the rover cannot cross) is denoted with the symbol "*".

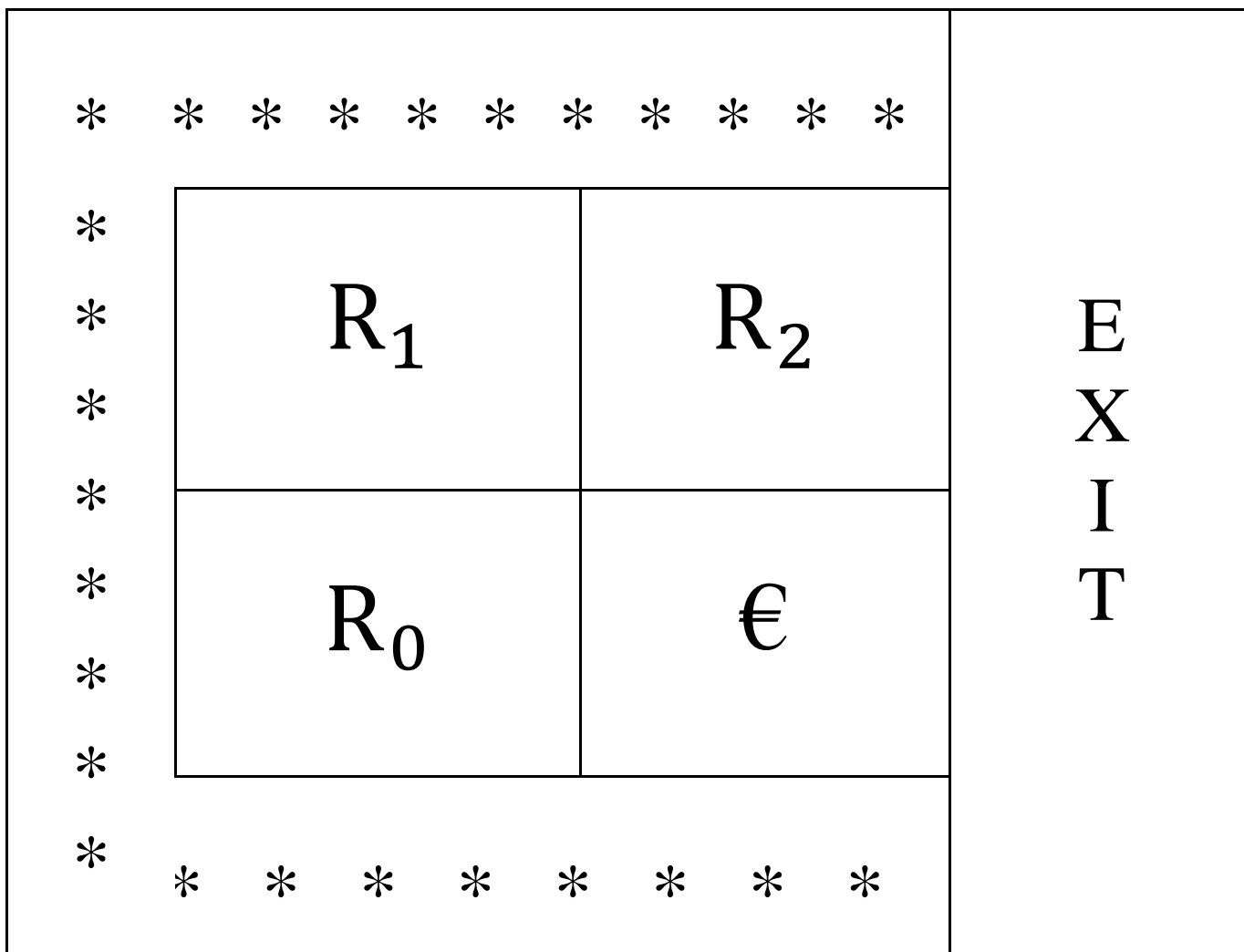

**Figure A.1.** The configuration of the *RockSample*(2,1) problem that we consider in our example. See the main text for illustration of the symbols.

The generative model used by the Active Inference agent is defined by setting the hidden states $S$, the observations $O$, the control states $U$, and the parameters $\Theta = \{\mathbf{A}, \mathbf{B}, \mathbf{C}, \mathbf{D}\}$.

As discussed in Section 4.3 of the main text, we decided not to factorize the hidden states $S$. Hence, with $n = 2$ and $k = 1$, the cardinality of $S$ is $(n^2 + 1) \cdot 2^k + 1 = 11$. See Fig. A.2. Note that each location corresponds to two states (e.g., the top-left location corresponds to states 4 and 5): one in which the only rock of the problem is bad (state 4) and the other in which the rock is good (state 5). The states 8, 9, and 10 are absorbing states; the former two states correspond to the EXIT location when the rock is bad (state 8) and when it is good (state 9), and the latter state corresponds to the border. The agent's initial belief is the vector $\mathbf{D} = [0,0,0.5,0.5,0,0,0,0,0,0,0]^{\mathrm{T}}$, which implies that it knows it starts from the top-left location, but it does not know whether the rock is bad or good (hence it considers equally probably starting from state 4 or 5).

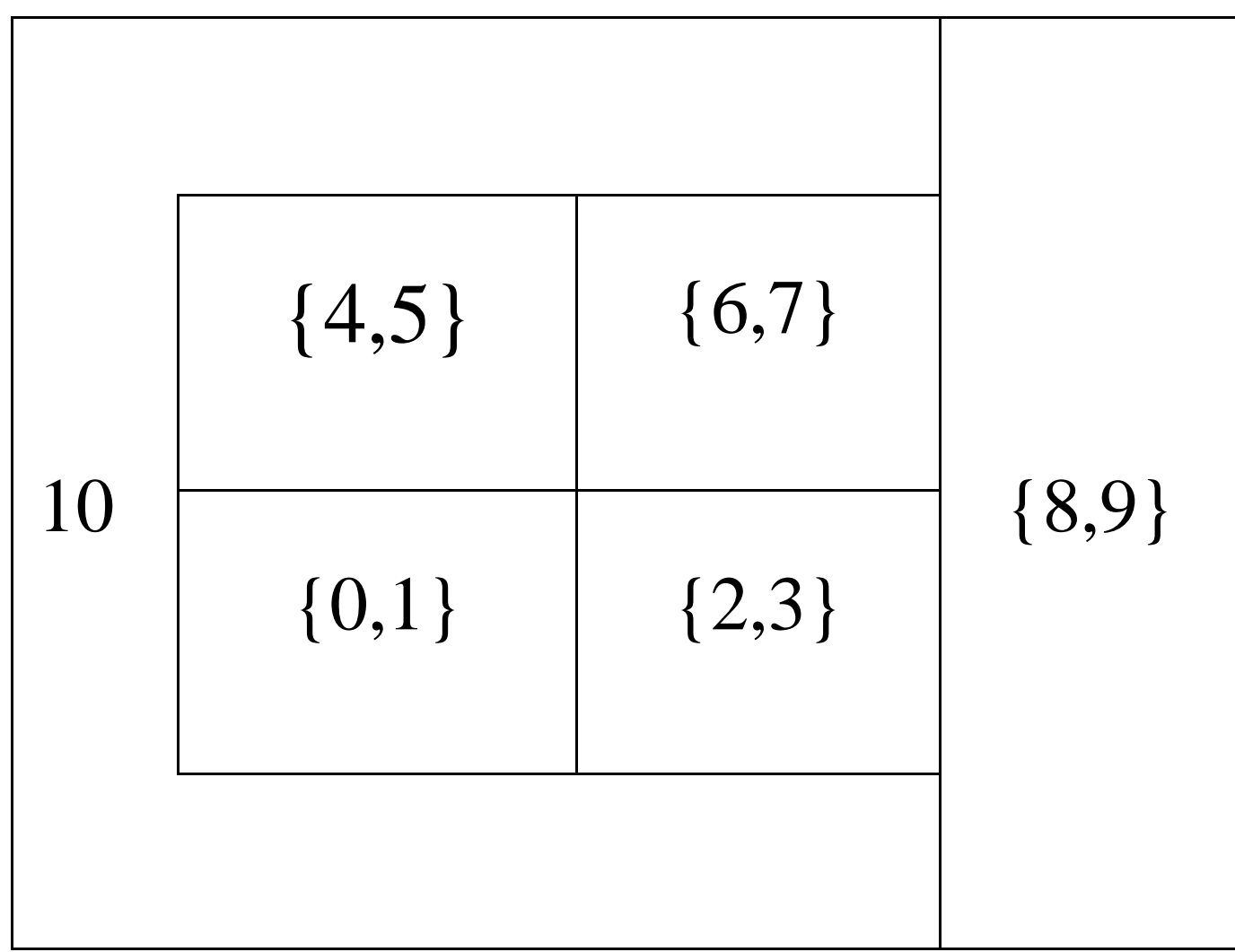

**Fig. A.2** Mapping between the 11 hidden states and the 6 grid locations of the *RockSample*(2,1) problem. See the main text for explanation.

Observations $O$ are organized in three factors, that is:

$$O \equiv \left[o_{R_0}, o_{€}, o_{R_1}, o_{R_2}, o_{\mathrm{EXIT}}, o_*\right]^T \otimes [\mathrm{bad, good}]^T \otimes [\mathrm{reward, penalty}]^T.$$

The first factor describes the 6 locations of the maze and therefore has size 6. The second factor describes all the possible $2^k$ possible combinations of rock qualities, good or bad (for $k = 1$, this factor has size 2). The third factor describes the utility (reward or penalty) of the current state. By considering that the main goal of the rover is obtaining rewards, the prior over (preferred) observations is:

$$\mathbf{C} = [1/6,1/6,1/6,1/6,1/6,1/6]^T \otimes [0.5,0.5]^T \otimes [\mathrm{c}, -\mathrm{c}]^T, \text{ with } c > 0.$$

The control states $U$ encode the actions of the agent. These are: go north (gn), go south (gs), go west (gw), go east (ge), check the rock from remote (cr) with variable accuracy, and collect a sample of the rock (sr). Note that for $k > 1$, there are $k$ additional remote sensing actions, one for each rock. Note also that the action cr changes the quality of the sampled rock from "good" to "bad", or leaves it unaltered if the rock was already bad.

The control states $U$ determine the transitions from one state to another, as specified by the set of matrices $\mathbf{B}_\mathrm{u}$ shown below, one for each control state $u$.

$$
B_{gn} = \begin{pmatrix}
0 & & & & & & & & & & \\
\vdots & & & & & & & & & & \\
 & & & & & & & & & & \\
0 & \vdots & & & & & & & & & \\
1 & 0 & \vdots & & & & & & & & \\
0 & 1 & 0 & \vdots & & & & & & & \\
\vdots & 0 & 1 & 0 & & & & & \vdots & & \\
 & \vdots & 0 & 1 & & & & & 0 & \vdots & \\
 & & \vdots & 0 & \vdots & \vdots & \vdots & \vdots & 1 & 0 & \vdots \\
 & & & 0 & 0 & 0 & 0 & 0 & 0 & 1 & 0 \\
 & & & 0 & 1 & 1 & 1 & 1 & 0 & 0 & 1
\end{pmatrix}
$$

$$
B_{gs} = \begin{pmatrix}
0 & \cdots & & 0 & 1 & 0 & \cdots & & & & \\
\vdots & & & & 0 & 1 & 0 & \cdots & & & \\
 & & & & \vdots & 0 & 1 & 0 & & & \\
 & & & & & \vdots & 0 & 1 & & & \\
 & & & & & & \vdots & 0 & & & \\
 & & & & & & & \vdots & & & \\
 & & & & & & & & \vdots & & \\
 & & & & & & & & 0 & \vdots & \\
\vdots & \vdots & \vdots & \vdots & & & & & 1 & 0 & \vdots \\
0 & 0 & 0 & 0 & \cdots & & & & 0 & 1 & 0 \\
1 & 1 & 1 & 1 & 0 & \cdots & & & \cdots & 0 & 1
\end{pmatrix}
$$

$$
B_{gw} = \begin{pmatrix}
0 & 0 & 1 & 0 & \cdots & & & & & & \\
\vdots & \vdots & 0 & 1 & & & & & & & \\
 & & \vdots & 0 & & & \vdots & & & & \\
 & & & \vdots & & & 0 & \vdots & & & \\
 & & & & & & 1 & 0 & & & \\
 & & & & & & 0 & 1 & & & \\
 & & & & & & \vdots & 0 & \vdots & & \\
 & & & & & & & \vdots & 0 & \vdots & \\
\vdots & \vdots & & & \vdots & \vdots & & & 1 & 0 & \vdots \\
0 & 0 & & & 0 & 0 & \vdots & \vdots & 0 & 1 & 0 \\
1 & 1 & & & 1 & 1 & 0 & 0 & 0 & 0 & 1
\end{pmatrix}
$$

$$
B_{ge} = \begin{pmatrix}
0 & & & & & & & & & & \\
0 & \vdots & & & & & & & & & \\
1 & 0 & & & & & & & & & \\
0 & 1 & & & & & & & & & \\
\vdots & 0 & & & \vdots & & & & & & \\
 & \vdots & & & 0 & \vdots & & & & & \\
 & & \vdots & & 1 & 0 & \vdots & & \vdots & & \\
 & & 0 & \vdots & 0 & 1 & 0 & \vdots & 0 & \vdots & \\
 & & 1 & 0 & \vdots & 0 & 1 & 0 & 1 & 0 & \vdots \\
 & & 0 & 1 & & \vdots & 0 & 1 & 0 & 1 & 0 \\
 & & 0 & 0 & & & 0 & 0 & 0 & 0 & 1
\end{pmatrix}
$$

$$
\mathrm{B_{cr}} = \begin{pmatrix}
1 & 0 & \cdots & & & & \\
0 & \ddots & & & & & \\
\vdots & & & & & & \\
& & \ddots & & & & \\
& & & 1 & & & \\
& & & & \ddots & & \\
& & & & & & \vdots \\
& & & & & \ddots & 0 \\
& & & & \cdots & 0 & 1
\end{pmatrix}
$$

$$
\mathrm{B_{sr}} = \begin{pmatrix}
1 & 0 & 0 & 0 & & & & & & & \\
0 & 1 & 0 & 0 & & & & & & & \\
\vdots & 0 & 1 & 1 & \vdots & & & & & & \\
& \vdots & 0 & 0 & 0 & \vdots & & & & & \\
& & \vdots & \vdots & 1 & 0 & \vdots & & & & \\
& & & & 0 & 1 & 0 & \vdots & & & \\
& & & & \vdots & 0 & 1 & 0 & \vdots & & \\
& & & & & \vdots & 0 & 1 & 0 & \vdots & \\
& & & & & & \vdots & 0 & 1 & 0 & \vdots \\
& & & & & & & \vdots & 0 & 1 & 0 \\
& & & & & & & & 0 & 0 & 1
\end{pmatrix}
$$

Furthermore, the control states $U$ determine the likelihood mapping from hidden states to observations, as specified in the matrix **A**. Note that the matrix **A** has three components, one for each factor of the observations. The first component $\mathrm{A_u^1}$, which describes the 6 locations of the maze, is the same for each $u$:

$$
\mathrm{A}_u^1 = \begin{pmatrix}
1 & 1 & 0 & 0 & 0 & 0 & & & & & \\
0 & 0 & 1 & 1 & 0 & 0 & \vdots & \vdots & & & \\
\vdots & \vdots & 0 & 0 & 1 & 1 & 0 & 0 & \vdots & \vdots & \\
& & \vdots & \vdots & 0 & 0 & 1 & 1 & 0 & 0 & \vdots \\
& & & & \vdots & \vdots & 0 & 0 & 1 & 1 & 0 \\
& & & & & & 0 & 0 & 0 & 0 & 1
\end{pmatrix}, \forall\, u \in U
$$

The second component $\mathrm{A}_u^2$, which describes the observed rock configuration, is different depending on the control state $u$ set. For $u$ expressing movements, namely $u \in \{\mathrm{gn}, \mathrm{gs}, \mathrm{gw}, \mathrm{ge}\}$, $\mathrm{A}_u^2$ is:

$$
\mathrm{A}^2_{\{\mathrm{gn,gs,gw,ge}\}} = \begin{pmatrix}
1 & 0 & 1 & 0 & 1 & 0 & 1 & 0 & 1 & 0 & 0.5 \\
0 & 1 & 0 & 1 & 0 & 1 & 0 & 1 & 0 & 1 & 0.5
\end{pmatrix}.
$$

For $u = \mathrm{cr}$, the $\mathrm{A}^2_{\mathrm{cr}}$ elements are computed by considering that the probability of observing the rock quality depends only on the rover position R and is independent of the other rock qualities. The probability that the sensor is accurate on the rock is $P_{ac|\mathrm{R}_i,\mathrm{cr}} \equiv P(\text{accurate} \mid \mathrm{R}_i\, , \mathrm{cr}) = \; (1 +$ $\eta(\mathrm{R}_i, €)) \,/\, 2$; where $\eta(\mathrm{R}_i, €) = 2^{-d(\mathrm{R}_i, €)/d_0}$, with $d(\mathrm{R}_i, €)$ denoting the Euclidean distance between the positions $\mathrm{R}_i$ and €, and $d_0$ is a constant specifying the half accuracy distance. In our case, there is only one rock, and its quality is "good". Therefore, $\mathrm{A}^2_{\mathrm{cr}}$ is:

$$\mathrm{A}^2_{\mathrm{cr}} = \begin{pmatrix} \bar{P}_{ac|\mathrm{R}_0,\mathrm{cr}} & \bar{P}_{ac|\mathrm{R}_0,\mathrm{cr}} & 0 & 0 & \bar{P}_{ac|\mathrm{R}_1,\mathrm{cr}} & \bar{P}_{ac|\mathrm{R}_1,\mathrm{cr}} & \bar{P}_{ac|\mathrm{R}_2,\mathrm{cr}} & \bar{P}_{ac|\mathrm{R}_2,\mathrm{cr}} & 1 & 0 & 0.5 \\ P_{ac|\mathrm{R}_0,\mathrm{cr}} & P_{ac|\mathrm{R}_0,\mathrm{cr}} & 1 & 1 & P_{ac|\mathrm{R}_1,\mathrm{cr}} & P_{ac|\mathrm{R}_1,\mathrm{cr}} & P_{ac|\mathrm{R}_2,\mathrm{cr}} & P_{ac|\mathrm{R}_2,\mathrm{cr}} & 0 & 1 & 0.5 \end{pmatrix}$$

where $\bar{P}_{ac|\mathrm{R}_i,\mathrm{cr}} = 1 - P_{ac|\mathrm{R}_i,\mathrm{cr}}$.

The third component $\mathrm{A}^3_u$, which describes the utility value associated with the observed outcome, is different depending on the control state $u$ set. The $\mathrm{A}^3_{\mathrm{sr}}$ matrix for the control state $u = \mathrm{sr}$ is the following:

$$\mathrm{A}^3_{\mathrm{sr}} = \begin{pmatrix} 0 & 0 & 0 & 1 & 0 & 0 & 0 & 0 & 0.5 & 0.5 & 0.5 \\ 1 & 1 & 1 & 0 & 1 & 1 & 1 & 1 & 0.5 & 0.5 & 0.5 \end{pmatrix}$$

The $\mathrm{A}^3_{\mathrm{sr}}$ matrix for all the other control states $u \in \{\mathrm{gn}, \mathrm{gs}, \mathrm{gw}, \mathrm{ge}, \mathrm{cr}\}$ is the following:

$$\mathrm{A}^3_{\{\mathrm{gn,gs,gw,ge,cr}\}} = \begin{pmatrix} 0.5 & \cdots & \cdots & 0.5 & 1 & 0 & 0 \\ 0.5 & \cdots & \cdots & 0.5 & 0 & 1 & 1 \end{pmatrix}$$

where the first and second rows encode the probability of observing a reward and a penalty, respectively.

The overall $\mathrm{A}^3_u$ matrix reflects the fact that the only two ways to observe a reward are collecting a sample of the good rock (i.e., being in state 3) and going to the EXIT after having collected a sample of the good rock (i.e., being in state 8, which implies that the rock is bad - which is possible after a sample of the good rock has been collected and the good rock has been changed in a bad rock). Rather, the agent gets a penalty if it reaches the EXIT without having collected a sample of the good rock, or if it reaches the border at any time.